\documentclass{article} 
\usepackage[table,xcdraw,dvipsnames]{xcolor}

\usepackage[]{iclr2024_conference,times}

\usepackage{amsmath,amsfonts,bm}

\def\eqref#1{equation~\ref{#1}}

\def\1{\bm{1}}

\DeclareMathAlphabet{\mathsfit}{\encodingdefault}{\sfdefault}{m}{sl}
\SetMathAlphabet{\mathsfit}{bold}{\encodingdefault}{\sfdefault}{bx}{n}

\newcommand{\set}[1]{\ensuremath{\mathcal{#1}}}

\usepackage{hyperref}
\usepackage{url}

\usepackage{graphicx}
\usepackage{float}

\definecolor{mygray}{gray}{0.5}
\definecolor{cblue}{RGB}{8, 85, 153}
\definecolor{darkblue}{RGB}{1, 43, 112}

\definecolor{cgreen}{RGB}{8, 153, 83}

\usepackage{amsmath}
\usepackage{amssymb}
\usepackage{amsthm}
\usepackage{booktabs}
\usepackage{longtable}
\usepackage{placeins} 
\usepackage{makecell}
\usepackage{multirow}

\usepackage[capitalize]{cleveref}
\usepackage{fontawesome}

\usepackage{adjustbox}

\usepackage{tabularx}

\newcommand{\method}{SASC}
\newcommand{\methods}{SASC }
\newcommand{\methodlongunderlineds}{\underline{S}ummarize \underline{a}nd \underline{Sc}ore }

\newcommand{\blank}{\underline{\hspace{15pt}}}

\theoremstyle{definition}

\crefname{definition}{Definition}{Definitions}%
\crefname{section}{Sec.}{Secs.}%

\usepackage{scrextend} 
\usepackage[noadjust]{cite}

\newcommand{\err}[1]{\scriptsize $\pm${#1}}
\usepackage{makecell}
\title{Explaining black box text modules\\in natural language with language models}

\author{
Chandan Singh$^{*, 1}$
\And
Aliyah R. Hsu$^{*, 1, 2}$
\And
Richard Antonello$^{3}$
\And
Shailee Jain$^{3}$
\And
Alexander G. Huth$^{3}$
\And
Bin Yu$^{1}$
\And
Jianfeng Gao$^{1}$
\And {} \And {} \And {}
\AND
\textnormal{$^1$ Microsoft Research}\;
\textnormal{$^2$ University of California, Berkeley}\\ 
\textnormal{$^3$ The University of Texas at Austin}\;
\textnormal{* Equal contribution}
}

\iclrfinalcopy 
\begin{document}

\maketitle

\begin{abstract}
    Large language models (LLMs) have demonstrated remarkable prediction performance for a growing array of tasks.
    However, their rapid proliferation and increasing opaqueness have created a growing need for interpretability.
    Here, we ask whether we can automatically obtain natural language explanations for black box text modules.
    A \textit{text module} is any function that maps text to a scalar continuous value, such as a submodule within an LLM or a fitted model of a brain region.
    \textit{Black box} indicates that we only have access to the module's inputs/outputs.
    
    We introduce \methodlongunderlineds (SASC), a method that takes in a text module and returns a natural language explanation of the module's selectivity along with a score for how reliable the explanation is.
    We study SASC in 3 contexts.
    First, we evaluate SASC on synthetic modules and find that it often recovers ground truth explanations.
    Second, we use SASC to explain modules found within a pre-trained BERT model, enabling inspection of the model's internals.
    Finally, we show that SASC can generate explanations for the response of individual fMRI voxels to language stimuli, with potential applications to fine-grained brain mapping.
    All code for using SASC and reproducing results is made available on Github.
    \footnote{Scikit-learn-compatible API available at \href{https://github.com/csinva/imodelsX}{\faGithub~github.com/csinva/imodelsX} and code for experiments along with all generated explanations is available at \href{https://github.com/microsoft/automated-explanations}{\faGithub~github.com/microsoft/automated-explanations}.}
\end{abstract}

\section{Introduction}

Large language models (LLMs) have demonstrated remarkable predictive performance across a growing range of diverse tasks~\citep{brown2020language,devlin2018bert}.
However, the inability to effectively interpret these models has led them to be characterized as black boxes.
This opaqueness has debilitated their use in high-stakes applications such as medicine~\citep{Kornblith2022},
and raised issues related to 
regulatory pressure \citep{goodman2016european}, safety~\citep{amodei2016concrete}, and alignment~\citep{gabriel2020artificial}.
This lack of interpretability is particularly detrimental in scientific fields, such as neuroscience~\citep{huth2016natural} or social science~\citep{ziems2023can}, where trustworthy interpretation itself is the end goal. 

To ameliorate these issues, we propose \methodlongunderlineds (\method).
\methods produces \textit{natural language explanations for text modules}.
We define a \textit{text module} $f$ as any function that maps text to a scalar continuous value, e.g. a neuron in a pre-trained LLM\footnote{Note that a neuron in an LLM typically returns a sequence-length vector rather than a scalar, so a transformation (e.g. averaging) is required before interpretation.}.
Given $f$, \methods returns a short natural language explanation describing what elicits the strongest response from $f$.
\methods requires only black-box access to the module (it does not require access to the module internals) and no human intervention.

\begin{figure}[t]
    \centering
    \vspace{-29pt}
    \includegraphics[width=0.9\textwidth]{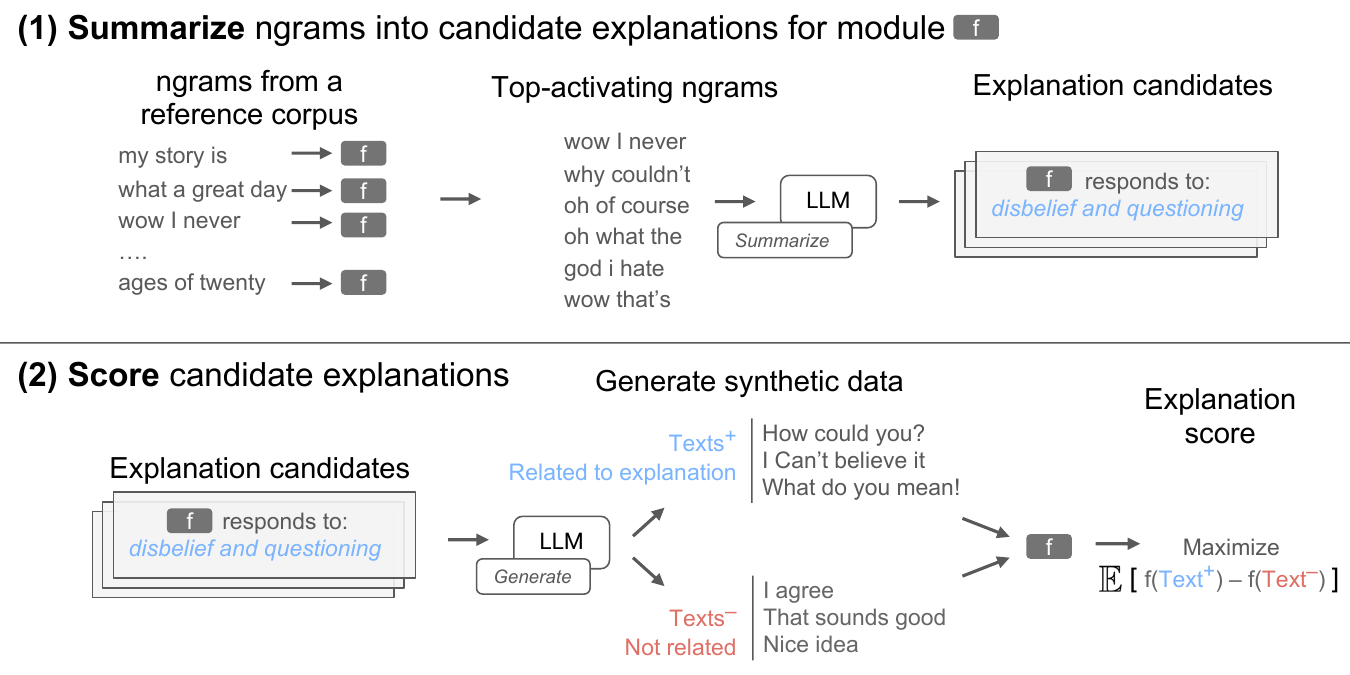}
    \vspace{-11pt}
    \caption{\methods pipeline for obtaining a natural language explanation given a module \textit{f}.
    \textbf{(i)} \methods first generates candidate explanations (using a pre-trained LLM)
    based on the ngrams that elicit the most positive response from $f$.
    \textbf{(ii)} \methods then evaluates each candidate explanation by generating synthetic data based on the explanation and testing the response of $f$ to the data.
    }
    \label{fig:intro}
\end{figure}

\method{} uses two steps to ground explanations in the responses of $f$ (\cref{fig:intro}).
In the first step, \methods derives explanation candidates by sorting $f$'s responses to ngrams and summarizing the top ngrams using a pre-trained LLM.
In the second step, \methods evaluates each candidate explanation by generating synthetic text based on the explanation (again with a pre-trained LLM) and testing the response of $f$ to the text;
these responses to synthetic text are used to assign an \textit{explanation score} to each explanation, that rates the reliability of the explanation.
Decomposing explanation into these separate steps helps mitigate issues with LLM hallucination when generating and evaluating explanations.

We evaluate \methods in two contexts.
In our main evaluation, we evaluate \methods on synthetic modules and find that it can often recover ground truth explanations under different experimental conditions (\cref{sec:synthetic_results}).
In our second evaluation, we use \methods to explain modules found within a pre-trained BERT model after applying dictionary learning (details in \cref{sec:bert_results}), and find that \methods explanations are often of comparable quality to human-given explanations (without the need for manual annotation).
Furthermore, we verify that BERT modules which are useful for downstream text-classification tasks often yield explanations related to the task.

The recovered explanations yield interesting insights.
Modules found within BERT respond to a variety of different phenomena, from individual words to broad, semantic concepts.
Additionally, we apply \methods to modules that are trained to predict the response of individual brain regions to language stimuli, as measured by fMRI.
We find that explanations for fMRI modules pertain more to social concepts (e.g. relationships and family) than BERT modules,
suggesting possible different emphases between modules in BERT and in the brain.
These explanations also provide fine-grained hypotheses about the selectivity of different brain regions to semantic concepts.

\section{Method}
\label{sec:methods}

\methods aims to interpret a text module $f$, which maps text to a scalar continuous value.
For example $f$ could be the output probability for a single token in an LLM,
or the output of a single neuron extracted from a vector of LLM activations.
\methods returns a short explanation describing what elicits the strongest response from $f$, along with an \textit{explanation score}, which rates how reliable the explanation is.
In the process of explanation, \methods uses a pre-trained \textit{helper LLM} to perform summarization and to generate synthetic text.
To mitigate potential hallucination introduced by the helper LLM, SASC decomposes the explanation process into 2 steps (\cref{fig:intro}) that greatly simplify the task performed by the helper LLM:

\paragraph{Step 1: Summarization}
The first step generates candidate explanations by summarizing ngrams.
All unique ngrams are extracted from a pre-specified corpus of text and fed through the module $f$.
The ngrams that elicit the largest positive response from $f$ are then fed through the helper LLM for summarization.
To avoid over-reliance on the very top ngrams, we select a random subset of the top ngrams in the summarization step.
This step is similar to prior works which summarize ngrams using manual inspection/parse trees~\citep{kadar2017representation,na2019discovery}, but the use of the helper LLM enables flexible, automated summarization.

The computational bottleneck of SASC is computing $f$'s response to the corpus ngrams.
This computation requires two choices:
the corpus underlying the extracted ngrams, and the length of ngrams to extract.
Using a larger corpus/higher order ngrams can make \methods more accurate, but the computational cost grows linearly with the unique number of ngrams in the corpus.
The corpus should be large enough to include relevant ngrams,
as the corpus limits what generated explanations are possible (e.g. it is difficult to recover mathematical explanations from a corpus that contains no math).
To speed up computation, ngrams can be subsampled from the corpus.

\paragraph{Step 2: Synthetic scoring}
The second step aims to evaluate each candidate explanation and select the most reliable one.
\methods generates synthetic data based on each candidate explanation, again using the helper LLM.
Intuitively, if the explanation accurately describes $f$, then $f$ should output large values for text related to the explanation (\textit{Text}$^+$) compared to unrelated synthetic text (\textit{Text}$^-$).\footnote{The unrelated synthetic text should be neutral text that omits the relevant explanation, but may introduce bias into the scoring if the helper LLM improperly generates negative synthetic texts. Instead of synthetic texts, a large set of neutral texts may be used for $\textit{Text}^-$, e.g. samples from a generic corpus.}
We then compute the explanation score as follows:
\begin{equation}
    \label{eq:score}
    \text{Explanation score} = \mathbb E [f(\textit{Text}^+) - f(\textit{Text}^-)] \text{ with units }\sigma_f,
\end{equation}

where a larger score corresponds to a more reliable explanation.
We report the score in units of $\sigma_f$, the standard deviation of $f$'s response to the corpus.
An explanation score of $1\sigma_f$ means that synthetic text related to the explanation increased the mean module response by one standard deviation compared to unrelated text.
\methods returns the candidate explanation that maximizes this difference, along with the synthetic data score.
The selection of the highest-scoring explanation is similar to the reranking step used in some prompting methods, e.g. \citep{shin2020autoprompt}, but differs in that it maximizes $f$'s response to synthetic data rather than optimizing the likelihood of a pre-specified dataset.

\paragraph{Limitations and hyperparameter settings}
While effective, the explanation pipeline described here has some clear limitations.
First and foremost, \methods assumes that $f$ can be concisely described in a natural language string.
This excludes complex functions or modules that respond to a non-coherent set of inputs.
Second, \methods only describes the inputs that elicit the largest responses from $f$, rather than its full behavior.
Finally, \methods requires that the pre-trained LLM can faithfully perform its required tasks (summarization and generation).
If an LLM is unable to perform these tasks sufficiently well, users may treat the output of SASC as candidate explanations to be vetted by a human, rather than final explanations to be used.

We use GPT-3 (\texttt{text-davinci-003}, Feb. 2023)~\citep{brown2020language} as the helper LLM (see LLM prompts in \cref{subsec:synthetic_appendix}).
In the summarization step, we use word-level trigrams, choose 30 random ngrams from the top 50 and generate 5 candidate explanations.
In the synthetic scoring step, we generate 20 synthetic strings (each is a sentence) for each candidate explanation, half of which are related to the  explanation.

\section{Recovering ground truth explanations for synthetic modules}
\label{sec:synthetic_results}

This section describes our main evaluation of \method: its ability to recover explanations for synthetic modules with a known ground truth explanation.

\paragraph{Experimental setup for synthetic modules}

We construct 54 synthetic modules based on the pre-trained Instructor embedding model~\citep{su2022one} (\texttt{hkunlp/instructor-xl}).
Each module is based on a dataset from a recent diverse collection~\citep{zhong2021adapting,zhong2022describing} that admits a simple, verifiable keyphrase description describing each underlying dataset, e.g. \textit{related to math} (full details in \cref{tab:synthetic_examples_full}).
Each module is constructed to return high values for text related to the module's groundtruth keyphrase and low values otherwise. Specifically, the module computes the Instructor embedding for an input text and for the groundtruth keyphrase; it then returns the negative Euclidean distance between the embeddings. We find that the synthetic modules reliably produce large values for text related to the desired keyphrase (\cref{fig:mean_preds_synthetic}).

We test \method's ability to recover accurate explanations for each of our 54 modules in 3 settings:
(1) The \textit{Default} setting extracts ngrams for summarization from the dataset corresponding to each module, which contains relevant ngrams for the ground truth explanation.
(2) The \textit{Restricted corpus} setting checks the impact of the underlying corpus on the performance of \method.
To do so, we restrict the ngrams we use for generating explanation candidates to a corpus from a random dataset among the 54, potentially containing less relevant ngrams.
(3) The \textit{Noisy module} setting adds Gaussian noise with standard deviation $3\sigma_f$ to all module responses in the summarization step.

\paragraph{Baselines and evaluation metrics}
We compare \method{} to three baselines:
(1) ngram-summarization, which summarizes top ngrams with an LLM, but does not use explanation scores to select among candidate explanations (essentially SASC without the scoring step);
(2) gradient-based explanations~\citep{poerner2018interpretable}, which use the gradients of $f$ with respect to the input to generate maximally activating inputs;
(3) topic modeling~\citep{blei2003latent}, which learns a 100-component dictionary over ngrams using latent dirichlet allocation.

We evaluate similarity of the recovered explanation and the groundtruth explanation in two ways:
(1) Accuracy: verifying whether the ground truth is essentially equivalent to the recovered explanation via manual inspection and
(2) BERT-score~\citep{zhang2019bertscore}\footnote{BERT-score is calculated with the base model \texttt{microsoft/deberta-xlarge-mnli}~\citep{he2021deberta}.}.
We find that these two metrics, when averaged over the datasets studied here, have a perfect rank correlation, i.e. every increase in average accuracy corresponds to an increase in average BERT score.
For topic modeling, accuracy is evaluated by taking the top-30 scoring ngrams for the module (as is done with SASC), finding the 5 topics with the highest scores for these ngrams, and manually checking whether there is a match between the groundtruth and any of the top-5 words in any of these topics.

\begin{table}[t]
    \centering
    \small
    \caption{Explanation recovery performance.
    For both metrics, higher is better.
    Each value is averaged over 54 modules and 3 random seeds; errors show standard error of the mean.}
    \begin{tabular}{lll|ll}
\toprule
& \multicolumn{2}{c}{\textbf{\method}} & \multicolumn{2}{c}{Baseline (ngram summarization)}\\
 & Accuracy & BERT Score & \makecell{Accuracy} & \makecell{BERT Score} \\
\midrule
Default & 0.883 \err{0.03} & 0.712 \err{0.02} & 0.753 \err{0.02} & 0.622 \err{0.05} \\
Restricted corpus & 0.667 \err{0.04} & 0.639 \err{0.02} & 0.540 \err{0.02} & 0.554 \err{0.05} \\
Noisy module & 0.679 \err{0.04} & 0.669 \err{0.02} & 0.456 \err{0.02} & 0.565 \err{0.06} \\
\midrule
Average & \textbf{0.743} & \textbf{0.673} & 0.582 & 0.580\\
\bottomrule
\end{tabular}

    \label{tab:recovery_results}

    \caption{Explanation recovery accuracy when varying hyperparameters for the \textit{Default} setting; averaged over 54 modules and 3 random seeds.}
    \small{
        \begin{tabular}{cccccc|cc}
    \toprule
    & \makecell{SASC\\(Original)} & \makecell{SASC\\(Bigrams)} & \makecell{SASC\\(4-grams)} & \makecell{SASC\\(LLaMA-2\\summarizer)} & \makecell{SASC\\(LLaMA-2\\generator)} & \makecell{Baseline\\(Gradient\\based)} & \makecell{Baseline\\(Topic\\modeling)} \\
    \midrule
    Acc. & 0.883\err{0.03} & 0.815\err{0.04} & 0.889\err{0.03} & 0.870\err{0.03} & 0.852\err{0.04} & 0.093\err{0.01} & 0.111\err{0.01} \\
    BERT Score & 0.712\err{0.02} & 0.690\err{0.03} & 0.714\err{0.02}& 0.705\err{0.02} & 0.701\err{0.02} & 0.351\err{0.01} &0.388\err{0.01} \\
    \bottomrule
    \end{tabular}
    }
    \label{tab:recovery_ablations}

    \small
    \caption{Examples of recovered explanations for different modules in the \textit{Default} setting.}
    \begin{tabular}{lll}
\toprule
  & Groundtruth Explanation & \methods Explanation \\
\midrule
\parbox[c]{1mm}{\multirow{6}{*}{\rotatebox[origin=c]{90} {Correct}}}
 & atheistic & atheism and related topics, such as theism, religious beliefs, and atheists \\
 & environmentalism & environmentalism and climate action \\
 & crime & crime and criminal activity \\
 & sports & sports \\
 & definition & defining or explaining something \\
 & facts & information or knowledge \\
\midrule
\parbox[c]{1mm}{\multirow{3}{*}{\rotatebox[origin=c]{90} {Incorrect}}} & derogatory & negative language and criticism \\
 & ungrammatical & language \\
 & subjective & art and expression \\
\bottomrule
\end{tabular}

    \label{tab:recovery_examples}
\end{table}

\paragraph{\methods can recover ground truth descriptions}
\cref{tab:recovery_results} shows the performance of \methods at recovering ground truth explanations.
In the \textit{Default} setting, \methods successfully identifies $88\%$ of the ground truth explanations.
In the two noisy settings, \methods still manages to recover explanations 67\% and 68\% of the time for the \textit{Restricted ngrams} and \textit{Noisy module} settings, respectively.
In all cases, \methods outperforms the ngram-summarization baseline.

\cref{tab:recovery_ablations} shows the results for the \textit{Default} setting when varying different modeling choices.
Performance is similar across various choices, such as using bigrams or 4-grams rather than trigrams in the summarization step, or when using the LLaMA-2 13-billion parameter model~\citep{touvron2023llama2} as the helper LLM rather than GPT-3.
Additionally, we find that explanation performance increases with the capabilities of the helper LLM used for summarization/generation (\cref{fig:synth_vary_llm}).
\cref{tab:recovery_ablations} also shows that the gradient-based baseline fails to accurately identify the underlying groundtruth text, consistent with previous work in prompting~\citep{singh2022explaining,shin2020autoprompt} and that topic modeling performs poorly, largely because the topic model fails to construct topics relevant to each specific module, as the same input ngrams are shared across all modules.

\cref{tab:recovery_examples} shows examples of correct and incorrect recovered explanations along with the ground truth explanation.
For some modules, \methods finds perfect keyword matches, e.g. \textit{sports}, or slight paraphrases, e.g. \textit{definition} $\to$ \textit{defining or explaining something}.
For the incorrect examples, the generated explanation is often similar to the ground truth explanation, e.g. \textit{derogatory} $\to$ \textit{negative language and criticism}, but occasionally, \methods fails to correctly identify the underlying pattern, e.g. \textit{ungrammatical} $\to$ \textit{language}.
Some failures may be due to the inability of ngrams to capture the underlying explanation, whereas others may be due to the constructed module imperfectly representing the ground truth explanation.

\begin{figure}[t]
    \centering
    \vspace{-22pt}
    \includegraphics[width=0.75\textwidth]{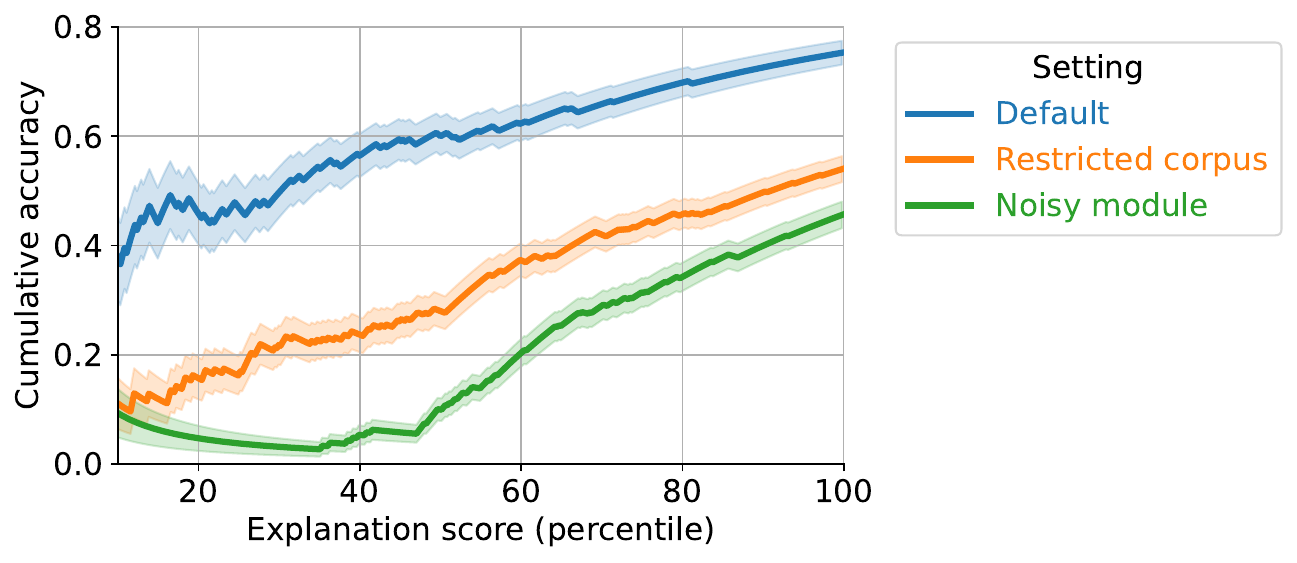}
    \vspace{-12pt}
    \caption{Cumulative accuracy at recovering the ground truth explanation increases as a function of explanation score.
    Error bars show standard error of the mean.
    }
    \label{fig:explanation_score_curves}
\end{figure}

\cref{fig:explanation_score_curves} shows the cumulative accuracy at recovering the ground truth explanation as a function of the explanation score.
Across all settings, accuracy increases as a function of explanation score, suggesting that higher explanation scores indicate more reliable explanations.
This also helps validate that the helper LLM is able to sucessfully generate useful synthetic texts for evaluation.

\section{Generating explanations for BERT transformer factors}
\label{sec:bert_results}

Next, we evaluate \methods using explanations for modules within BERT~\citep{devlin2018bert} (\texttt{bert-base-uncased}).
In the absence of ground truth explanations, we evaluated the explanations by (i) comparing them to human-given explanations and (ii) checking their relevance to downstream tasks.

\paragraph{BERT transformer factor modules}
One can interpret any module within BERT, e.g. a single neuron or an expert in an MOE~\citep{fedus2022review};
here, we choose to interpret \textit{transformer factors}, following a previous study that suggests that they are amenable to interpretation~\citep{yun2021transformer}.
Transformer factors learn a transformation of activations across layers via dictionary learning (details in \cref{subsec:transformer_factors_appendix}; corpus used is the WikiText dataset~\citep{merity2016pointer}).
Each transformer factor is a module that takes as input a text sequence and yields a scalar dictionary coefficient, after averaging over the input's sequence length.
There are 1,500 factors, and their coefficients vary for each of BERT's 13 encoding layers.

\paragraph{Comparison to human-given explanations}
\cref{tab:bert_examples} compares \methods explanations to those given by humans in prior work  (31 unique explanations from Table 1, Table 3 and Appendix in \citep{yun2021transformer}).
They are sometimes similar with different phrasings,
e.g. \textit{leaving or being left} versus \textit{Word ``left''}, 
and sometimes quite different, e.g. \textit{publishing, media, or awards} versus \textit{Institution with abbreviation.}
For each transformer factor, we compare the explanation scores for \methods and the human-given explanations.
The \methods explanation score is higher 61\% of the time and \method's mean explanation score is $1.6\sigma_f$ compared to $1.0\sigma_f$ for the human explanation.
This evaluation
suggests that the \methods explanations can be of similar quality to the human explanations, despite requiring no manual effort.

\begin{table}[t]
    \centering
    \caption{Comparing sample \methods to human-labeled explanations for BERT transformer factors.
    Win percentage shows how often the SASC explanation yields a higher explanation score than the human explanation.
    See all explanations and scores in \cref{tab:bert_examples_full}.}
    \small
    \renewcommand{\arraystretch}{1.2}
    \makebox[\textwidth][c]{
    \begin{tabular}{ll p{0.49\textwidth} p{0.49\textwidth}}
\toprule

\methods Explanation                                                                           & Human Explanation                                                                              \\
\midrule
names of parks                                                                                                    & Word “park”. Noun. a common first and last name.                                                             \\
leaving or being left                                                                                             & Word “left". Verb. leaving, exiting                                                                          \\
specific dates or months                                                                                          & Consecutive years, used in football season naming.                                                           \\
idea of wrongdoing or illegal activity                                                                            & something unfortunate happened.                                                                              \\
introduction of something new                                                                                     & Doing something again, or making something new again.                                                        \\
versions or translations                                                                                          & repetitive structure detector.                                                                               \\
publishing, media, or awards                                                                                      & Institution with abbreviation.                                                                               \\
names of places, people, or things                                                                                & Unit exchange with parentheses                                                                               \\
\midrule
\methods win percentage: \textbf{61\%} & Human explanation win percentage: 39\%\\
\methods mean explanation score: \textbf{1.6$\mathbf \sigma_f$} & Human explanation mean explanation score: 1.0$\sigma_f$\\
\bottomrule
\end{tabular}
    }
    \label{tab:bert_examples}
\end{table}

\paragraph{Mapping explained modules to text-classification tasks}
We now investigate whether the learned \methods explanations are useful for informing which downstream tasks a module is useful for.
Given a classification dataset where the input $X$ is a list of $n$ strings and the output $y$ is a list of $n$ class labels, we first convert $X$ to a matrix of transformer factor coefficients $X_{TF} \in \mathbb{R}^{n \times 19,500}$, where each row contains the concatenated factor coefficients across layers.
We then fit a sparse logistic regression model to $(X_{TF}, y)$, and analyze the explanations for the factors with the 25 largest coefficients across all classes.
Ideally, these explanations would be relevant to the text-classification task; 
we evaluate what fraction of the 25 explanations are relevant for each task via manual inspection.

We study 3 widely used text-classification datasets: \textit{emotion}~\citep{saravia2018carer} (classifying tweet emotion as sadness, joy, love, anger, fear or surprise),
\textit{ag-news}~\citep{Zhang2015CharacterlevelCN} (classifying news headlines as world, sports, business, or sci/tech),
and 
\textit{SST2} \citep{socher2013recursive} (classifying movie review sentiment as positive or negative).
\cref{tab:bert_linear} shows results evaluating the BERT transformer factor modules selected by a sparse linear model fit to these datasets.
A large fraction of the explanations for selected modules are, in fact, relevant to their usage in downstream tasks, ranging from 0.35 for \textit{Emotion} to 0.96 for \textit{AG News}.
The \textit{AG News} task has a particularly large fraction of relevant explanations, with many explanations corresponding very directly to class labels, e.g. \textit{professional sports teams} $\to$ \textit{sports} or \textit{financial investments} $\to$ \textit{business}.
See the full set of generated explanations in \cref{subsec:transformer_factors_appendix}.

\begin{table}[t]
    \centering
    \footnotesize
    \caption{BERT modules selected by a sparse linear model fit to text-classification tasks.
    First row shows the fraction of explanations for the selected modules which are relevant to the downstream task.
    Second row shows test accuracy for the fitted linear models.
    Bottom section shows sample explanations for modules selected by the linear model which are relevant to the downstream task.
    Values are averaged over 3 random linear model fits (error bars show the standard error of the mean).
    }
    \renewcommand{\arraystretch}{1.3}
    \makebox[\textwidth][c]{
    \begin{tabular}{p{0.08 \textwidth} p{0.25 \textwidth} p{0.3\textwidth} p{0.25\textwidth}}
\toprule
         & Emotion & AG News & SST2\\
         \midrule
         \makecell[l]{Fraction\\relevant} & 0.35\err{0.082} & 0.96\err{0.033} & 0.44\err{0.086}\\
         \makecell[l]{Test\\accuracy} & 0.75\err{0.001} & 0.81\err{0.001} & 0.84\err{0.001}\\
         \midrule
         \parbox[c]{1mm}{\multirow{6}{*}{\rotatebox[origin=c]{90}{Sample relevant}}}\hspace{3pt}
         \parbox[c]{1mm}{\multirow{6}{*}{\rotatebox[origin=c]{90}{explanations}}}
          & negative emotions such as hatred, disgust, disdain, rage, and horror & people, places, or things related to japan & a negative statement, usually in the form of not or nor\\
         & injury or impairment & professional sports teams & hatred and violence\\
         & humor & geography & harm, injury, or damage\\
         & romance & financial investments & something being incorrect or wrong\\
         \bottomrule
\end{tabular}
    }
    \label{tab:bert_linear}
\end{table}

\begin{figure}[t]
    \centering
    \vspace{-8pt}
    \includegraphics[width=0.75\textwidth]{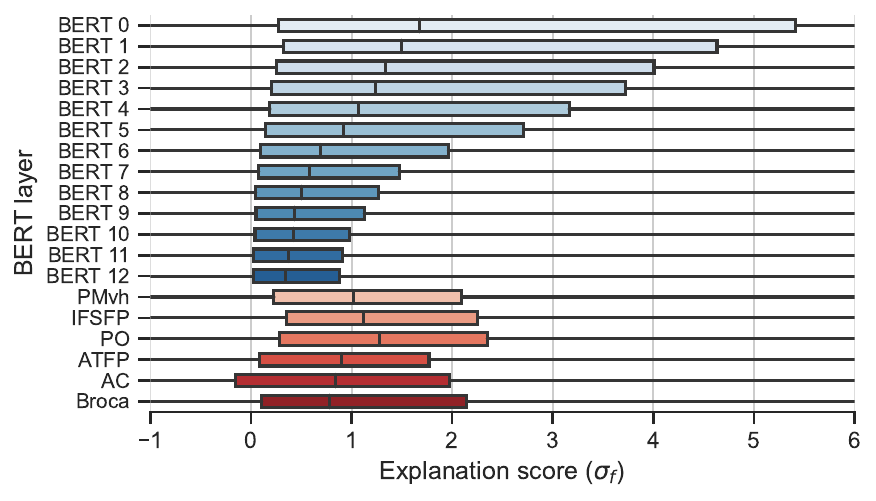}
    \vspace{-12pt}
    \caption{Explanation score for BERT (blue) and fMRI (orange) modules.
    As the BERT layer increases, the explanation score tends to decrease, implying modules are harder to explain with \method.
    Across regions, explanation scores for fMRI voxel modules are generally lower than scores for BERT modules in early layers and comparable to scores for the final layers.
    Boxes show the median and interquartile range.
    ROI abbreviations: premotor ventral hand area (PMvh), 
    anterior temporal face patch (ATFP),
    auditory cortex (AC),
    parietal operculum (PO),
    inferior frontal sulcus face patch (IFSFP),
    Broca's area (Broca).
    }
    \label{fig:syn_perc_score_boxplot}

    \centering
    \includegraphics[width=0.9\textwidth]{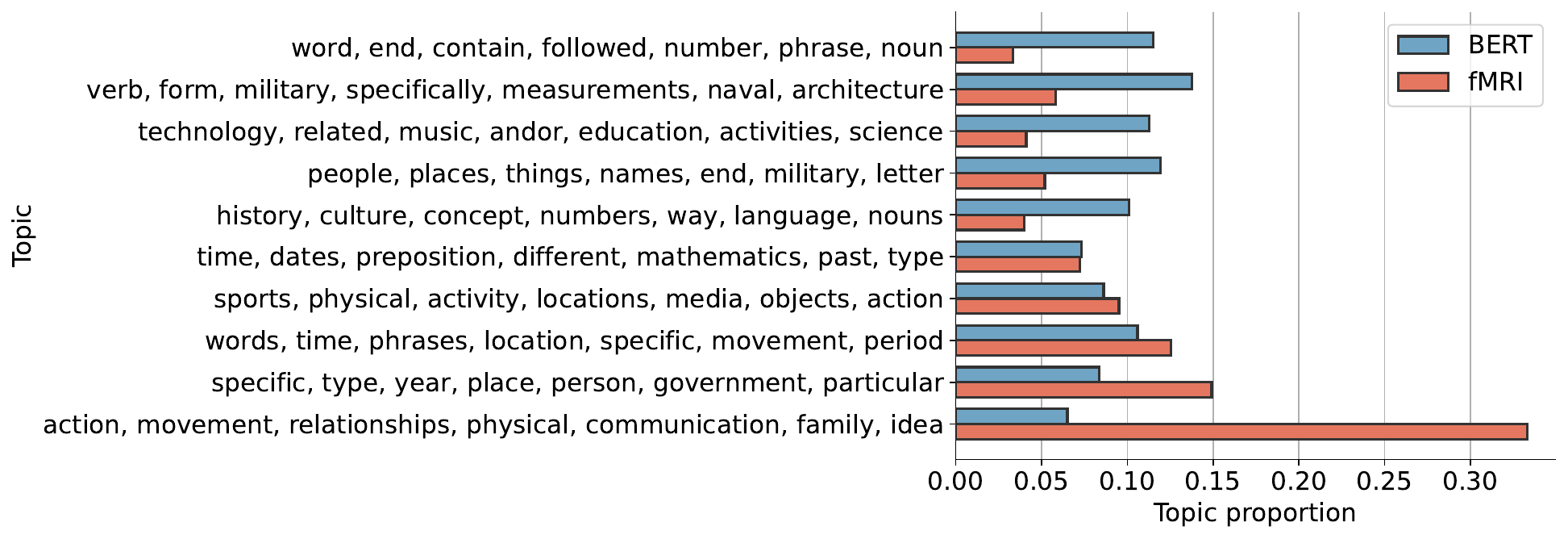}
    \vspace{-8pt}
    \caption{Topics found by LDA for explanations of BERT factors and fMRI voxels.
    Topic proportion is calculated by assigning each explanation to the topic with the largest coefficient.
    Topic proportions for BERT/fMRI explanations largely overlap, although the bottom topic consisting of physical/social words is much more prevalent in fMRI explanations.}
    \label{fig:topic_proportions}
\end{figure}

\paragraph{Patterns in \methods explanations}
\methods provides 1,500 explanations for transformer factors in 13 layers of BERT.
\cref{fig:syn_perc_score_boxplot} shows that the explanation score decreases with increasing layer depth, suggesting that \methods better explains factors at lower layers.
The mean explanation score across all layers is 1.77$\sigma_f$.

To understand the breakdown of topics present in the explanations,
we fit a topic model (with Latent Dirichlet Allocation~\citep{blei2003latent}) to the remaining explanations.
The topic model has 10 topics and preprocesses each explanation by converting it to a vector of word counts.
We exclude all factors that do not attain an explanation score of at least $1\sigma_f$ from the topic model, as they are less likely to be correct.
\cref{fig:topic_proportions} shows each topic along with the proportion of modules whose largest topic coefficient is for that topic.
Topics span a wide range of categories, from syntactic concepts (e.g. \textit{word, end, ..., noun}) to more semantic concepts (e.g. \textit{sports, physical, activity, ...}).

\section{Generating explanations for fMRI-voxel modules}
\label{sec:fmri_results}

\paragraph{fMRI voxel modules}
A central challenge in neuroscience is understanding how and where semantic concepts are represented in the brain.
To meet this challenge, one line of study predicts the response of different brain voxels (i.e. small regions in the brain) to natural language stimuli~\citep{huth2016natural,jain2018incorporating}.
We analyze data from \citep{lebel2022natural} and \citep{tang2023semantic}, which consists of fMRI responses for 3 human subjects as they listen to 20+ hours of narrative stories from podcasts.
We fit modules\ to predict the fMRI response in each voxel from the text that the subject was hearing by extracting text embeddings with a pre-trained LLaMA model (\texttt{decapoda-research/llama-30b-hf})~\citep{touvron2023llama}.
After fitting the modules on the training split and evaluating them on the test split using bootstrapped ridge regression, we generate \methods explanations for 1,500 well-predicted voxel modules, distributed evenly among the three human subjects and diverse cortical areas (see details on the fMRI experimental setup in \cref{subsec:fmri_setup}).

\paragraph{Voxel explanations}
\cref{tab:fmri_explanations} shows examples of explanations for individual voxels, along with three top ngrams used to derive the explanation.
Each explanation unifies fairly different ngrams under a common theme, e.g. \textit{sliced cucumber, cut the apples, sauteed shiitake...} $\to$ \textit{food preparation}.
In some cases, the explanations recover language concepts similar to known selectivity in sensory modalities, e.g. face selectivity in IFSFP ~\citep{tsao_patches_2008} and selectivity for non-speech sounds such as laughter in primary auditory cortex~\citep{hamilton2021parallel}.
The ngrams also provide more fine-grained hypotheses for selectivity (e.g. \textit{physical injury or pain}) compared to the coarse semantic categories proposed in earlier language studies (e.g. \textit{emotion}~\citep{huth2016natural, binder_where_2009, mitchell_predicting_2008}).

\cref{fig:topic_proportions} shows the topics that fMRI explanations best fit into compared with BERT transformer factors.
The proportions for many topics are similar, but the fMRI explanations yield a much greater proportion for the topic consisting of social words (e.g. \textit{relationships}, \textit{communication}, \textit{family}) and perceptual words (e.g. \textit{action}, \textit{movement}, \textit{physical}). This is consistent with prior knowledge, as the largest axis of variation for fMRI voxels is known to separate social concepts from physical concepts~\citep{huth2016natural}.

The selected 1,500 voxels often achieve explanation scores considerably greater than zero for their explanations (mean explanation score $1.27 \sigma_f \pm 0.029$).
\cref{fig:syn_perc_score_boxplot} (bottom) shows the mean explanation score for the six most common fMRI regions of interest (ROIs) among the voxels we study here.
Across regions, the fMRI voxel modules generally attain explanation scores that are slightly lower than BERT modules in early layers and slightly higher than BERT modules in the final layers.
We also find some evidence that the generated fMRI voxel explanations can explain not just the fitted module, but also brain responses to unseen data (see \cref{subsec:fmri_voxel_eval_supp}).
This suggests that the voxel explanations here can serve as hypotheses for followup experiments to affirm the fine-grained selectivity of specific brain voxels.


\section{Related work}
\label{sec:related_work}

\begin{table}[H]
    \centering
    \footnotesize
    \caption{
    Examples of recovered explanations for individual fMRI voxel modules.
    All achieve an fMRI predicted correlation greater than 0.3 and an explanation score of at least $1\sigma$.
    The third column shows 3 of the ngrams used to derive the explanation in the \methods summarization step.
    }
    \makebox[\textwidth][c]{
    \begin{tabular}{lll}
\toprule
Explanation & ROI & Example top ngrams \\
\midrule
looking or staring in some way  & IFSFP & eyed her suspiciously, wink at, locks eyes with\\
relationships and loss  & ATFP  & girlfriend now ex, lost my husband, was a miscarriage \\
physical injury or pain & Broca & infections and gangrene, pulled a muscle, burned the skin\\
counting or measuring time  & PMvh & count down and, weeks became months, three more seconds\\
food preparation  & ATFP & sliced cucumber, cut the apples, sauteed shiitake\\
laughter or amusement & ATFP, AC & started to laugh, funny guy, chuckled and\\
\bottomrule
\end{tabular}

    }
    \label{tab:fmri_explanations}
\end{table}

\paragraph{Explaining modules in natural language}
A few related works study generating natural language explanations.
MILAN~\citep{hernandez2022natural} uses patch-level information of visual features to generate descriptions of neuron behavior in vision models.
iPrompt~\citep{singh2022explaining} uses automated prompt engineering and D5~\citep{zhong2023goaldd,zhong2022describing}/GSClip~\citep{zhu2022gsclip} use LLMs to describe patterns in a dataset (as opposed to describing a module, as we study here).
In concurrent work, \citep{bills2023language} propose an algorithm similar to \methods that explains individual neurons in an LLM by predicting token-level neuron activations.

Two very related works use top-activating ngrams/sentences to construct explanations:
(1) \citep{kadar2017representation} builds an explanation by \textit{manually} inspecting the top ngrams eliciting the largest module responses from a corpus using an omission-based approach.
(2) \citep{na2019discovery} similarly extracts the top sentences from a corpus, but summarizes them using a parse tree.
Alternatively, \citep{poerner2018interpretable} use a gradient-based method to generate maximally activating text inputs.

\paragraph{Explaining neural-network predictions}
Most prior works have focused on the problem of explaining a \emph{single prediction} with natural language, rather than an entire module, e.g. for text classification~\citep{camburu2018snli,rajani2019explain,narang2020wt5}, 
or computer vision~\citep{hendricks2016generating,zellers2019recognition}.
Besides natural language explanations, 
some works explain individual prediction via
feature importances~(e.g. LIME~\citep{ribeiro2016model}/SHAP~\citep{lundberg2019explainable}),
feature-interaction importances~\citep{morris2023tree,singh2019Hierarchical,tsang2017detecting},
or extractive rationales~\citep{zaidan2008modeling,sha2021learning}.
They are not directly comparable to \method, as they work at the prediction-level and do not produce a natural-language explanation.

\paragraph{Explaining neural-network representations}
We build on a long line of recent work that explains neural-network \textit{representations},
e.g. via probing~\citep{conneau2018you,liu2019incorporating},
via visualization~\citep{zeiler2014visualizing,karpathy2015visualizing},
by categorizing neurons into categories~\citep{bau2017network,bau2018gan,bau2020understanding,dalvi2019one,gurnee2023finding},
localizing knowledge in an LLM~\citep{meng2022locating,dai2021knowledge},
or distilling information into a transparent model \citep{tan2018distill,ha2021adaptive,singh2023augmenting}.

\paragraph{Natural language representations in fMRI}

Using the representations from LLMs to help predict brain responses to natural language has become common among neuroscientists studying language processing in recent years~\citep{jain2018incorporating,wehbe_aligning_2014,schrimpf2021neural,TONEVANEURIPS2019,antonello2021low,goldstein_shared_2022}.
This paradigm of using ``encoding models'' \citep{wu_complete_2006} to better understand how the brain processes language has been applied to help understand the cortical organization of language timescales \citep{JAINNEURIPS2020, chen2023cortical}, examine the relationship between visual and semantic information in the brain \citep{popham2021visual}, and explore to what extent syntax, semantics or discourse drives brain activity \citep{caucheteux_disentangling_2021, kauf2023lexical, reddy_can_2020, pasquiou_information-restricted_2023, aw_training_2022, kumar_reconstructing_2022, oota_joint_2022,tuckute2023driving}.

\section{Discussion}
\label{sec:discussion}

\methods could potentially enable much better mechanistic interpretability for LLMs, allowing for automated analysis of submodules present in LLMs (e.g. attention heads, transformer factors, or experts in an MOE),
along with an explanation score that helps inform when an explanation is reliable.
Trustworthy explanations could help audit increasingly powerful LLMs for undesired behavior or improve the distillation of smaller task-specific modules.
\methods also could also be a useful tool in many scientific pipelines.
The fMRI analysis performed here generates many explanations which can be directly tested via followup fMRI experiments to understand the fine-grained selectivity of brain regions.
\methods could also be used to generate explanations in a variety of domains, such as analysis of text models in computational social science or in medicine.

While effective, \methods has many limitations.
\methods only explains a module's top responses, but it could be extended to explain the entirety of the module's responses (e.g. by selecting top ngrams differently).
Additionally, due to its reliance on ngrams, SASC fails to capture low-level text patterns or patterns requiring long context, e.g. patterns based on position in a sequence.
Future explanations could consider adding information beyond ngrams, and also probe the relationships between different modules to explain circuits of modules rather than modules in isolation.

{
    \footnotesize
    \bibliography{main.bib}

\begin{thebibliography}{86}
\providecommand{\natexlab}[1]{#1}
\providecommand{\url}[1]{\texttt{#1}}
\expandafter\ifx\csname urlstyle\endcsname\relax
  \providecommand{\doi}[1]{doi: #1}\else
  \providecommand{\doi}{doi: \begingroup \urlstyle{rm}\Url}\fi

\bibitem[Amodei et~al.(2016)Amodei, Olah, Steinhardt, Christiano, Schulman, and
  Man{\'e}]{amodei2016concrete}
Dario Amodei, Chris Olah, Jacob Steinhardt, Paul Christiano, John Schulman, and
  Dan Man{\'e}.
\newblock Concrete problems in ai safety.
\newblock \emph{arXiv preprint arXiv:1606.06565}, 2016.

\bibitem[Antonello et~al.(2021)Antonello, Turek, Vo, and
  Huth]{antonello2021low}
Richard Antonello, Javier~S Turek, Vy~Vo, and Alexander Huth.
\newblock Low-dimensional structure in the space of language representations is
  reflected in brain responses.
\newblock \emph{Advances in Neural Information Processing Systems},
  34:\penalty0 8332--8344, 2021.

\bibitem[Aw \& Toneva(2022)Aw and Toneva]{aw_training_2022}
Khai~Loong Aw and Mariya Toneva.
\newblock Training language models for deeper understanding improves brain
  alignment, December 2022.
\newblock URL \url{http://arxiv.org/abs/2212.10898}.
\newblock arXiv:2212.10898 [cs, q-bio].

\bibitem[Bau et~al.(2017)Bau, Zhou, Khosla, Oliva, and
  Torralba]{bau2017network}
David Bau, Bolei Zhou, Aditya Khosla, Aude Oliva, and Antonio Torralba.
\newblock Network dissection: Quantifying interpretability of deep visual
  representations.
\newblock In \emph{Proceedings of the IEEE conference on computer vision and
  pattern recognition}, pp.\  6541--6549, 2017.

\bibitem[Bau et~al.(2018)Bau, Zhu, Strobelt, Zhou, Tenenbaum, Freeman, and
  Torralba]{bau2018gan}
David Bau, Jun-Yan Zhu, Hendrik Strobelt, Bolei Zhou, Joshua~B Tenenbaum,
  William~T Freeman, and Antonio Torralba.
\newblock Gan dissection: Visualizing and understanding generative adversarial
  networks.
\newblock \emph{arXiv preprint arXiv:1811.10597}, 2018.

\bibitem[Bau et~al.(2020)Bau, Zhu, Strobelt, Lapedriza, Zhou, and
  Torralba]{bau2020understanding}
David Bau, Jun-Yan Zhu, Hendrik Strobelt, Agata Lapedriza, Bolei Zhou, and
  Antonio Torralba.
\newblock Understanding the role of individual units in a deep neural network.
\newblock \emph{Proceedings of the National Academy of Sciences}, 117\penalty0
  (48):\penalty0 30071--30078, 2020.

\bibitem[Bills et~al.(2023)Bills, Cammarata, Mossing, Saunders, Wu, Tillman,
  Gao, Goh, Sutskever, and Leike]{bills2023language}
Steven Bills, Nick Cammarata, Dan Mossing, William Saunders, Jeff Wu, Henk
  Tillman, Leo Gao, Gabriel Goh, Ilya Sutskever, and Jan Leike.
\newblock Language models can explain neurons in language models.
\newblock
  \url{https://openaipublic.blob.core.windows.net/neuron-explainer/paper/index.html},
  2023.

\bibitem[Binder et~al.(2009)Binder, Desai, Graves, and
  Conant]{binder_where_2009}
Jeffrey~R. Binder, Rutvik~H. Desai, William~W. Graves, and Lisa~L. Conant.
\newblock Where {Is} the {Semantic} {System}? {A} {Critical} {Review} and
  {Meta}-{Analysis} of 120 {Functional} {Neuroimaging} {Studies}.
\newblock \emph{Cerebral Cortex}, 19\penalty0 (12):\penalty0 2767--2796,
  December 2009.
\newblock ISSN 1460-2199, 1047-3211.
\newblock \doi{10.1093/cercor/bhp055}.
\newblock URL
  \url{https://academic.oup.com/cercor/article-lookup/doi/10.1093/cercor/bhp055}.

\bibitem[Blei et~al.(2003)Blei, Ng, and Jordan]{blei2003latent}
David~M Blei, Andrew~Y Ng, and Michael~I Jordan.
\newblock Latent dirichlet allocation.
\newblock \emph{Journal of machine Learning research}, 3\penalty0
  (Jan):\penalty0 993--1022, 2003.

\bibitem[Brown et~al.(2020)Brown, Mann, Ryder, Subbiah, Kaplan, Dhariwal,
  Neelakantan, Shyam, Sastry, Askell, et~al.]{brown2020language}
Tom Brown, Benjamin Mann, Nick Ryder, Melanie Subbiah, Jared~D Kaplan, Prafulla
  Dhariwal, Arvind Neelakantan, Pranav Shyam, Girish Sastry, Amanda Askell,
  et~al.
\newblock Language models are few-shot learners.
\newblock \emph{Advances in neural information processing systems},
  33:\penalty0 1877--1901, 2020.

\bibitem[Camburu et~al.(2018)Camburu, Rockt{\"a}schel, Lukasiewicz, and
  Blunsom]{camburu2018snli}
Oana-Maria Camburu, Tim Rockt{\"a}schel, Thomas Lukasiewicz, and Phil Blunsom.
\newblock e-snli: Natural language inference with natural language
  explanations.
\newblock \emph{Advances in Neural Information Processing Systems}, 31, 2018.

\bibitem[Caucheteux et~al.(2021)Caucheteux, Gramfort, and
  King]{caucheteux_disentangling_2021}
Charlotte Caucheteux, Alexandre Gramfort, and Jean-Remi King.
\newblock Disentangling syntax and semantics in the brain with deep networks.
\newblock In \emph{Proceedings of the 38th {International} {Conference} on
  {Machine} {Learning}}, pp.\  1336--1348. PMLR, July 2021.
\newblock URL \url{https://proceedings.mlr.press/v139/caucheteux21a.html}.
\newblock ISSN: 2640-3498.

\bibitem[Caucheteux et~al.(2022)Caucheteux, Gramfort, and
  King]{caucheteux2022deep}
Charlotte Caucheteux, Alexandre Gramfort, and Jean-R{\'e}mi King.
\newblock Deep language algorithms predict semantic comprehension from brain
  activity.
\newblock \emph{Scientific Reports}, 12\penalty0 (1):\penalty0 16327, 2022.

\bibitem[Chen et~al.(2023)Chen, Dupr{\'e}~la Tour, Gallant, Klein, and
  Deniz]{chen2023cortical}
Catherine Chen, Tom Dupr{\'e}~la Tour, Jack Gallant, Daniel Klein, and Fatma
  Deniz.
\newblock The cortical representation of language timescales is shared between
  reading and listening.
\newblock \emph{bioRxiv}, pp.\  2023--01, 2023.

\bibitem[Conneau et~al.(2018)Conneau, Kruszewski, Lample, Barrault, and
  Baroni]{conneau2018you}
Alexis Conneau, German Kruszewski, Guillaume Lample, Lo{\"\i}c Barrault, and
  Marco Baroni.
\newblock What you can cram into a single vector: Probing sentence embeddings
  for linguistic properties.
\newblock \emph{arXiv preprint arXiv:1805.01070}, 2018.

\bibitem[Dai et~al.(2021)Dai, Dong, Hao, Sui, Chang, and Wei]{dai2021knowledge}
Damai Dai, Li~Dong, Yaru Hao, Zhifang Sui, Baobao Chang, and Furu Wei.
\newblock Knowledge neurons in pretrained transformers.
\newblock \emph{arXiv preprint arXiv:2104.08696}, 2021.

\bibitem[Dalvi et~al.(2019)Dalvi, Durrani, Sajjad, Belinkov, Bau, and
  Glass]{dalvi2019one}
Fahim Dalvi, Nadir Durrani, Hassan Sajjad, Yonatan Belinkov, Anthony Bau, and
  James Glass.
\newblock What is one grain of sand in the desert? analyzing individual neurons
  in deep nlp models.
\newblock In \emph{Proceedings of the AAAI Conference on Artificial
  Intelligence}, volume~33, pp.\  6309--6317, 2019.

\bibitem[Devlin et~al.(2018)Devlin, Chang, Lee, and Toutanova]{devlin2018bert}
Jacob Devlin, Ming-Wei Chang, Kenton Lee, and Kristina Toutanova.
\newblock Bert: Pre-training of deep bidirectional transformers for language
  understanding.
\newblock \emph{arXiv preprint arXiv:1810.04805}, 2018.

\bibitem[Fedus et~al.(2022)Fedus, Dean, and Zoph]{fedus2022review}
William Fedus, Jeff Dean, and Barret Zoph.
\newblock A review of sparse expert models in deep learning.
\newblock \emph{arXiv preprint arXiv:2209.01667}, 2022.

\bibitem[Fischl(2012)]{fischl2012freesurfer}
Bruce Fischl.
\newblock Freesurfer.
\newblock \emph{Neuroimage}, 62\penalty0 (2):\penalty0 774--781, 2012.

\bibitem[Gabriel(2020)]{gabriel2020artificial}
Iason Gabriel.
\newblock Artificial intelligence, values, and alignment.
\newblock \emph{Minds and machines}, 30\penalty0 (3):\penalty0 411--437, 2020.

\bibitem[Goldstein et~al.(2022)Goldstein, Zada, Buchnik, Schain, Price, Aubrey,
  Nastase, Feder, Emanuel, Cohen, Jansen, Gazula, Choe, Rao, Kim, Casto, Fanda,
  Doyle, Friedman, Dugan, Melloni, Reichart, Devore, Flinker, Hasenfratz, Levy,
  Hassidim, Brenner, Matias, Norman, Devinsky, and
  Hasson]{goldstein_shared_2022}
Ariel Goldstein, Zaid Zada, Eliav Buchnik, Mariano Schain, Amy Price, Bobbi
  Aubrey, Samuel~A. Nastase, Amir Feder, Dotan Emanuel, Alon Cohen, Aren
  Jansen, Harshvardhan Gazula, Gina Choe, Aditi Rao, Catherine Kim, Colton
  Casto, Lora Fanda, Werner Doyle, Daniel Friedman, Patricia Dugan, Lucia
  Melloni, Roi Reichart, Sasha Devore, Adeen Flinker, Liat Hasenfratz, Omer
  Levy, Avinatan Hassidim, Michael Brenner, Yossi Matias, Kenneth~A. Norman,
  Orrin Devinsky, and Uri Hasson.
\newblock Shared computational principles for language processing in humans and
  deep language models.
\newblock \emph{Nature Neuroscience}, 25\penalty0 (3):\penalty0 369--380, March
  2022.
\newblock ISSN 1546-1726.
\newblock \doi{10.1038/s41593-022-01026-4}.
\newblock URL \url{https://www.nature.com/articles/s41593-022-01026-4}.
\newblock Number: 3 Publisher: Nature Publishing Group.

\bibitem[Goodman \& Flaxman(2016)Goodman and Flaxman]{goodman2016european}
Bryce Goodman and Seth Flaxman.
\newblock European union regulations on algorithmic decision-making and a"
  right to explanation".
\newblock \emph{arXiv preprint arXiv:1606.08813}, 2016.

\bibitem[Gurnee et~al.(2023)Gurnee, Nanda, Pauly, Harvey, Troitskii, and
  Bertsimas]{gurnee2023finding}
Wes Gurnee, Neel Nanda, Matthew Pauly, Katherine Harvey, Dmitrii Troitskii, and
  Dimitris Bertsimas.
\newblock Finding neurons in a haystack: Case studies with sparse probing,
  2023.

\bibitem[Ha et~al.(2021)Ha, Singh, Lanusse, Upadhyayula, and
  Yu]{ha2021adaptive}
Wooseok Ha, Chandan Singh, Francois Lanusse, Srigokul Upadhyayula, and Bin Yu.
\newblock Adaptive wavelet distillation from neural networks through
  interpretations.
\newblock \emph{Advances in Neural Information Processing Systems}, 34, 2021.

\bibitem[Hamilton et~al.(2021)Hamilton, Oganian, Hall, and
  Chang]{hamilton2021parallel}
Liberty~S Hamilton, Yulia Oganian, Jeffery Hall, and Edward~F Chang.
\newblock Parallel and distributed encoding of speech across human auditory
  cortex.
\newblock \emph{Cell}, 184\penalty0 (18):\penalty0 4626--4639, 2021.

\bibitem[He et~al.(2021)He, Liu, Gao, and Chen]{he2021deberta}
Pengcheng He, Xiaodong Liu, Jianfeng Gao, and Weizhu Chen.
\newblock Deberta: Decoding-enhanced bert with disentangled attention.
\newblock In \emph{International Conference on Learning Representations}, 2021.
\newblock URL \url{https://openreview.net/forum?id=XPZIaotutsD}.

\bibitem[Hendricks et~al.(2016)Hendricks, Akata, Rohrbach, Donahue, Schiele,
  and Darrell]{hendricks2016generating}
Lisa~Anne Hendricks, Zeynep Akata, Marcus Rohrbach, Jeff Donahue, Bernt
  Schiele, and Trevor Darrell.
\newblock Generating visual explanations.
\newblock In \emph{European conference on computer vision}, pp.\  3--19.
  Springer, 2016.

\bibitem[Hernandez et~al.(2022)Hernandez, Schwettmann, Bau, Bagashvili,
  Torralba, and Andreas]{hernandez2022natural}
Evan Hernandez, Sarah Schwettmann, David Bau, Teona Bagashvili, Antonio
  Torralba, and Jacob Andreas.
\newblock Natural language descriptions of deep visual features.
\newblock In \emph{International Conference on Learning Representations}, 2022.

\bibitem[Huth et~al.(2016)Huth, De~Heer, Griffiths, Theunissen, and
  Gallant]{huth2016natural}
Alexander~G Huth, Wendy~A De~Heer, Thomas~L Griffiths, Fr{\'e}d{\'e}ric~E
  Theunissen, and Jack~L Gallant.
\newblock Natural speech reveals the semantic maps that tile human cerebral
  cortex.
\newblock \emph{Nature}, 532\penalty0 (7600):\penalty0 453--458, 2016.

\bibitem[Jain \& Huth(2018)Jain and Huth]{jain2018incorporating}
Shailee Jain and Alexander Huth.
\newblock Incorporating context into language encoding models for fmri.
\newblock \emph{Advances in neural information processing systems}, 31, 2018.

\bibitem[Jain et~al.(2020)Jain, Vo, Mahto, LeBel, Turek, and
  Huth]{JAINNEURIPS2020}
Shailee Jain, Vy~Vo, Shivangi Mahto, Amanda LeBel, Javier~S Turek, and
  Alexander Huth.
\newblock Interpretable multi-timescale models for predicting fmri responses to
  continuous natural speech.
\newblock In H.~Larochelle, M.~Ranzato, R.~Hadsell, M.~F. Balcan, and H.~Lin
  (eds.), \emph{Advances in Neural Information Processing Systems}, volume~33,
  pp.\  13738--13749. Curran Associates, Inc., 2020.
\newblock URL
  \url{https://proceedings.neurips.cc/paper/2020/file/9e9a30b74c49d07d8150c8c83b1ccf07-Paper.pdf}.

\bibitem[K{\'a}d{\'a}r et~al.(2017)K{\'a}d{\'a}r, Chrupa{\l}a, and
  Alishahi]{kadar2017representation}
Akos K{\'a}d{\'a}r, Grzegorz Chrupa{\l}a, and Afra Alishahi.
\newblock Representation of linguistic form and function in recurrent neural
  networks.
\newblock \emph{Computational Linguistics}, 43\penalty0 (4):\penalty0 761--780,
  2017.

\bibitem[Karpathy et~al.(2015)Karpathy, Johnson, and
  Fei-Fei]{karpathy2015visualizing}
Andrej Karpathy, Justin Johnson, and Li~Fei-Fei.
\newblock Visualizing and understanding recurrent networks.
\newblock \emph{arXiv preprint arXiv:1506.02078}, 2015.

\bibitem[Kauf et~al.(2023)Kauf, Tuckute, Levy, Andreas, and
  Fedorenko]{kauf2023lexical}
Carina Kauf, Greta Tuckute, Roger Levy, Jacob Andreas, and Evelina Fedorenko.
\newblock Lexical semantic content, not syntactic structure, is the main
  contributor to ann-brain similarity of fmri responses in the language
  network.
\newblock \emph{bioRxiv}, pp.\  2023--05, 2023.

\bibitem[Kornblith et~al.(2022)Kornblith, Singh, Devlin, Addo, Streck, Holmes,
  Kuppermann, Grupp-Phelan, Fineman, Butte, and Yu]{Kornblith2022}
Aaron~E. Kornblith, Chandan Singh, Gabriel Devlin, Newton Addo, Christian~J.
  Streck, James~F. Holmes, Nathan Kuppermann, Jacqueline Grupp-Phelan, Jeffrey
  Fineman, Atul~J. Butte, and Bin Yu.
\newblock Predictability and stability testing to assess clinical decision
  instrument performance for children after blunt torso trauma.
\newblock \emph{PLOS Digital Health}, 2022.
\newblock \doi{https://doi.org/10.1371/journal.pdig.0000076}.
\newblock URL
  \url{https://journals.plos.org/digitalhealth/article?id=10.1371/journal.pdig.0000076}.

\bibitem[Kumar et~al.(2022)Kumar, Sumers, Yamakoshi, Goldstein, Hasson, Norman,
  Griffiths, Hawkins, and Nastase]{kumar_reconstructing_2022}
Sreejan Kumar, Theodore~R. Sumers, Takateru Yamakoshi, Ariel Goldstein, Uri
  Hasson, Kenneth~A. Norman, Thomas~L. Griffiths, Robert~D. Hawkins, and
  Samuel~A. Nastase.
\newblock Reconstructing the cascade of language processing in the brain using
  the internal computations of a transformer-based language model.
\newblock Technical report, bioRxiv, June 2022.
\newblock URL
  \url{https://www.biorxiv.org/content/10.1101/2022.06.08.495348v1}.
\newblock Section: New Results Type: article.

\bibitem[LeBel et~al.(2022)LeBel, Wagner, Jain, Adhikari-Desai, Gupta,
  Morgenthal, Tang, Xu, and Huth]{lebel2022natural}
Amanda LeBel, Lauren Wagner, Shailee Jain, Aneesh Adhikari-Desai, Bhavin Gupta,
  Allyson Morgenthal, Jerry Tang, Lixiang Xu, and Alexander~G Huth.
\newblock A natural language fmri dataset for voxelwise encoding models.
\newblock \emph{bioRxiv}, pp.\  2022--09, 2022.

\bibitem[Liu \& Avci(2019)Liu and Avci]{liu2019incorporating}
Frederick Liu and Besim Avci.
\newblock Incorporating priors with feature attribution on text classification.
\newblock \emph{arXiv preprint arXiv:1906.08286}, 2019.

\bibitem[Lundberg et~al.(2019)Lundberg, Erion, Chen, DeGrave, Prutkin, Nair,
  Katz, Himmelfarb, Bansal, and Lee]{lundberg2019explainable}
Scott~M Lundberg, Gabriel Erion, Hugh Chen, Alex DeGrave, Jordan~M Prutkin,
  Bala Nair, Ronit Katz, Jonathan Himmelfarb, Nisha Bansal, and Su-In Lee.
\newblock Explainable ai for trees: From local explanations to global
  understanding.
\newblock \emph{arXiv preprint arXiv:1905.04610}, 2019.

\bibitem[Meng et~al.(2022)Meng, Bau, Andonian, and Belinkov]{meng2022locating}
Kevin Meng, David Bau, Alex Andonian, and Yonatan Belinkov.
\newblock Locating and editing factual knowledge in gpt.
\newblock \emph{arXiv preprint arXiv:2202.05262}, 2022.

\bibitem[Merity et~al.(2016)Merity, Xiong, Bradbury, and
  Socher]{merity2016pointer}
Stephen Merity, Caiming Xiong, James Bradbury, and Richard Socher.
\newblock Pointer sentinel mixture models, 2016.

\bibitem[Mitchell et~al.(2008)Mitchell, Shinkareva, Carlson, Chang, Malave,
  Mason, and Just]{mitchell_predicting_2008}
Tom~M. Mitchell, Svetlana~V. Shinkareva, Andrew Carlson, Kai-Min Chang,
  Vicente~L. Malave, Robert~A. Mason, and Marcel~Adam Just.
\newblock Predicting human brain activity associated with the meanings of
  nouns.
\newblock \emph{Science (New York, N.Y.)}, 320\penalty0 (5880):\penalty0
  1191--1195, May 2008.
\newblock ISSN 1095-9203.
\newblock \doi{10.1126/science.1152876}.

\bibitem[Morris et~al.(2023)Morris, Singh, Rush, Gao, and Deng]{morris2023tree}
John~X Morris, Chandan Singh, Alexander~M Rush, Jianfeng Gao, and Yuntian Deng.
\newblock Tree prompting: Efficient task adaptation without fine-tuning.
\newblock \emph{arXiv preprint arXiv:2310.14034}, 2023.

\bibitem[Na et~al.(2019)Na, Choe, Lee, and Kim]{na2019discovery}
Seil Na, Yo~Joong Choe, Dong-Hyun Lee, and Gunhee Kim.
\newblock Discovery of natural language concepts in individual units of cnns.
\newblock \emph{arXiv preprint arXiv:1902.07249}, 2019.

\bibitem[Narang et~al.(2020)Narang, Raffel, Lee, Roberts, Fiedel, and
  Malkan]{narang2020wt5}
Sharan Narang, Colin Raffel, Katherine Lee, Adam Roberts, Noah Fiedel, and
  Karishma Malkan.
\newblock Wt5?! training text-to-text models to explain their predictions.
\newblock \emph{arXiv preprint arXiv:2004.14546}, 2020.

\bibitem[Nishimoto et~al.(2017)Nishimoto, Huth, Bilenko, and
  Gallant]{nishimoto2017eye}
Shinji Nishimoto, Alexander~G Huth, Natalia~Y Bilenko, and Jack~L Gallant.
\newblock Eye movement-invariant representations in the human visual system.
\newblock \emph{Journal of vision}, 17\penalty0 (1):\penalty0 11--11, 2017.

\bibitem[Oota et~al.(2022)Oota, Gupta, and Toneva]{oota_joint_2022}
Subba~Reddy Oota, Manish Gupta, and Mariya Toneva.
\newblock Joint processing of linguistic properties in brains and language
  models, December 2022.
\newblock URL \url{http://arxiv.org/abs/2212.08094}.
\newblock arXiv:2212.08094 [cs, q-bio].

\bibitem[Pasquiou et~al.(2023)Pasquiou, Lakretz, Thirion, and
  Pallier]{pasquiou_information-restricted_2023}
Alexandre Pasquiou, Yair Lakretz, Bertrand Thirion, and Christophe Pallier.
\newblock Information-{Restricted} {Neural} {Language} {Models} {Reveal}
  {Different} {Brain} {Regions}' {Sensitivity} to {Semantics}, {Syntax} and
  {Context}, February 2023.
\newblock URL \url{http://arxiv.org/abs/2302.14389}.
\newblock arXiv:2302.14389 [cs].

\bibitem[Poerner et~al.(2018)Poerner, Roth, and
  Sch{\"u}tze]{poerner2018interpretable}
Nina Poerner, Benjamin Roth, and Hinrich Sch{\"u}tze.
\newblock Interpretable textual neuron representations for nlp.
\newblock \emph{arXiv preprint arXiv:1809.07291}, 2018.

\bibitem[Popham et~al.(2021)Popham, Huth, Bilenko, Deniz, Gao, Nunez-Elizalde,
  and Gallant]{popham2021visual}
Sara~F Popham, Alexander~G Huth, Natalia~Y Bilenko, Fatma Deniz, James~S Gao,
  Anwar~O Nunez-Elizalde, and Jack~L Gallant.
\newblock Visual and linguistic semantic representations are aligned at the
  border of human visual cortex.
\newblock \emph{Nature neuroscience}, 24\penalty0 (11):\penalty0 1628--1636,
  2021.

\bibitem[Radford et~al.(2019)Radford, Wu, Child, Luan, Amodei, Sutskever,
  et~al.]{radford2019language}
Alec Radford, Jeffrey Wu, Rewon Child, David Luan, Dario Amodei, Ilya
  Sutskever, et~al.
\newblock Language models are unsupervised multitask learners.
\newblock \emph{OpenAI blog}, 1\penalty0 (8):\penalty0 9, 2019.

\bibitem[Rajani et~al.(2019)Rajani, McCann, Xiong, and
  Socher]{rajani2019explain}
Nazneen~Fatema Rajani, Bryan McCann, Caiming Xiong, and Richard Socher.
\newblock Explain yourself! leveraging language models for commonsense
  reasoning.
\newblock \emph{arXiv preprint arXiv:1906.02361}, 2019.

\bibitem[Reddy \& Wehbe(2020)Reddy and Wehbe]{reddy_can_2020}
Aniketh~Janardhan Reddy and Leila Wehbe.
\newblock Can {fMRI} reveal the representation of syntactic structure in the
  brain?
\newblock preprint, Neuroscience, June 2020.
\newblock URL \url{http://biorxiv.org/lookup/doi/10.1101/2020.06.16.155499}.

\bibitem[Ribeiro et~al.(2016)Ribeiro, Singh, and Guestrin]{ribeiro2016model}
Marco~Tulio Ribeiro, Sameer Singh, and Carlos Guestrin.
\newblock Model-agnostic interpretability of machine learning.
\newblock \emph{arXiv preprint arXiv:1606.05386}, 2016.

\bibitem[Saravia et~al.(2018)Saravia, Liu, Huang, Wu, and
  Chen]{saravia2018carer}
Elvis Saravia, Hsien-Chi~Toby Liu, Yen-Hao Huang, Junlin Wu, and Yi-Shin Chen.
\newblock Carer: Contextualized affect representations for emotion recognition.
\newblock In \emph{Proceedings of the 2018 conference on empirical methods in
  natural language processing}, pp.\  3687--3697, 2018.

\bibitem[Schrimpf et~al.(2021)Schrimpf, Blank, Tuckute, Kauf, Hosseini,
  Kanwisher, Tenenbaum, and Fedorenko]{schrimpf2021neural}
Martin Schrimpf, Idan~Asher Blank, Greta Tuckute, Carina Kauf, Eghbal~A
  Hosseini, Nancy Kanwisher, Joshua~B Tenenbaum, and Evelina Fedorenko.
\newblock The neural architecture of language: Integrative modeling converges
  on predictive processing.
\newblock \emph{Proceedings of the National Academy of Sciences}, 118\penalty0
  (45):\penalty0 e2105646118, 2021.

\bibitem[Sha et~al.(2021)Sha, Camburu, and Lukasiewicz]{sha2021learning}
Lei Sha, Oana-Maria Camburu, and Thomas Lukasiewicz.
\newblock Learning from the best: Rationalizing predictions by adversarial
  information calibration.
\newblock In \emph{AAAI}, pp.\  13771--13779, 2021.

\bibitem[Shin et~al.(2020)Shin, Razeghi, Logan~IV, Wallace, and
  Singh]{shin2020autoprompt}
Taylor Shin, Yasaman Razeghi, Robert~L Logan~IV, Eric Wallace, and Sameer
  Singh.
\newblock Autoprompt: Eliciting knowledge from language models with
  automatically generated prompts.
\newblock \emph{arXiv preprint arXiv:2010.15980}, 2020.

\bibitem[Singh et~al.(2019)Singh, Murdoch, and Yu]{singh2019Hierarchical}
Chandan Singh, W~James Murdoch, and Bin Yu.
\newblock Hierarchical interpretations for neural network predictions.
\newblock \emph{International Conference on Learning Representations}, pp.\
  ~26, 2019.
\newblock URL \url{https://openreview.net/forum?id=SkEqro0ctQ}.

\bibitem[Singh et~al.(2022{\natexlab{a}})Singh, Askari, Caruana, and
  Gao]{singh2023augmenting}
Chandan Singh, Armin Askari, Rich Caruana, and Jianfeng Gao.
\newblock Augmenting interpretable models with llms during training.
\newblock \emph{arXiv preprint arXiv:2209.11799}, 2022{\natexlab{a}}.

\bibitem[Singh et~al.(2022{\natexlab{b}})Singh, Morris, Aneja, Rush, and
  Gao]{singh2022explaining}
Chandan Singh, John~X Morris, Jyoti Aneja, Alexander~M Rush, and Jianfeng Gao.
\newblock Explaining patterns in data with language models via interpretable
  autoprompting.
\newblock \emph{arXiv preprint arXiv:2210.01848}, 2022{\natexlab{b}}.

\bibitem[Socher et~al.(2013)Socher, Perelygin, Wu, Chuang, Manning, Ng, and
  Potts]{socher2013recursive}
Richard Socher, Alex Perelygin, Jean Wu, Jason Chuang, Christopher~D Manning,
  Andrew Ng, and Christopher Potts.
\newblock Recursive deep models for semantic compositionality over a sentiment
  treebank.
\newblock In \emph{Proceedings of the 2013 conference on empirical methods in
  natural language processing}, pp.\  1631--1642, 2013.

\bibitem[Su et~al.(2022)Su, Kasai, Wang, Hu, Ostendorf, Yih, Smith,
  Zettlemoyer, Yu, et~al.]{su2022one}
Hongjin Su, Jungo Kasai, Yizhong Wang, Yushi Hu, Mari Ostendorf, Wen-tau Yih,
  Noah~A Smith, Luke Zettlemoyer, Tao Yu, et~al.
\newblock One embedder, any task: Instruction-finetuned text embeddings.
\newblock \emph{arXiv preprint arXiv:2212.09741}, 2022.

\bibitem[Tan et~al.(2018)Tan, Caruana, Hooker, and Lou]{tan2018distill}
Sarah Tan, Rich Caruana, Giles Hooker, and Yin Lou.
\newblock Distill-and-compare: Auditing black-box models using transparent
  model distillation.
\newblock In \emph{Proceedings of the 2018 AAAI/ACM Conference on AI, Ethics,
  and Society}, pp.\  303--310, 2018.

\bibitem[Tang et~al.(2023)Tang, LeBel, Jain, and Huth]{tang2023semantic}
Jerry Tang, Amanda LeBel, Shailee Jain, and Alexander~G Huth.
\newblock Semantic reconstruction of continuous language from non-invasive
  brain recordings.
\newblock \emph{Nature Neuroscience}, pp.\  1--9, 2023.

\bibitem[Toneva \& Wehbe(2019)Toneva and Wehbe]{TONEVANEURIPS2019}
Mariya Toneva and Leila Wehbe.
\newblock Interpreting and improving natural-language processing (in machines)
  with natural language-processing (in the brain).
\newblock In H.~Wallach, H.~Larochelle, A.~Beygelzimer, F.~d\textquotesingle
  Alch\'{e}-Buc, E.~Fox, and R.~Garnett (eds.), \emph{Advances in Neural
  Information Processing Systems}, volume~32. Curran Associates, Inc., 2019.
\newblock URL
  \url{https://proceedings.neurips.cc/paper/2019/file/749a8e6c231831ef7756db230b4359c8-Paper.pdf}.

\bibitem[Touvron et~al.(2023{\natexlab{a}})Touvron, Lavril, Izacard, Martinet,
  Lachaux, Lacroix, Rozi{\`e}re, Goyal, Hambro, Azhar,
  et~al.]{touvron2023llama}
Hugo Touvron, Thibaut Lavril, Gautier Izacard, Xavier Martinet, Marie-Anne
  Lachaux, Timoth{\'e}e Lacroix, Baptiste Rozi{\`e}re, Naman Goyal, Eric
  Hambro, Faisal Azhar, et~al.
\newblock Llama: Open and efficient foundation language models.
\newblock \emph{arXiv preprint arXiv:2302.13971}, 2023{\natexlab{a}}.

\bibitem[Touvron et~al.(2023{\natexlab{b}})Touvron, Martin, Stone, Albert,
  Almahairi, Babaei, Bashlykov, Batra, Bhargava, Bhosale,
  et~al.]{touvron2023llama2}
Hugo Touvron, Louis Martin, Kevin Stone, Peter Albert, Amjad Almahairi, Yasmine
  Babaei, Nikolay Bashlykov, Soumya Batra, Prajjwal Bhargava, Shruti Bhosale,
  et~al.
\newblock Llama 2: Open foundation and fine-tuned chat models.
\newblock \emph{arXiv preprint arXiv:2307.09288}, 2023{\natexlab{b}}.

\bibitem[Tsang et~al.(2017)Tsang, Cheng, and Liu]{tsang2017detecting}
Michael Tsang, Dehua Cheng, and Yan Liu.
\newblock Detecting statistical interactions from neural network weights.
\newblock \emph{arXiv preprint arXiv:1705.04977}, 2017.

\bibitem[Tsao et~al.(2008)Tsao, Schweers, Moeller, and
  Freiwald]{tsao_patches_2008}
Doris~Y. Tsao, Nicole Schweers, Sebastian Moeller, and Winrich~A. Freiwald.
\newblock Patches of face-selective cortex in the macaque frontal lobe.
\newblock \emph{Nature Neuroscience}, 11\penalty0 (8):\penalty0 877--879,
  August 2008.
\newblock ISSN 1546-1726.
\newblock \doi{10.1038/nn.2158}.

\bibitem[Tuckute et~al.(2023)Tuckute, Sathe, Srikant, Taliaferro, Wang,
  Schrimpf, Kay, and Fedorenko]{tuckute2023driving}
Greta Tuckute, Aalok Sathe, Shashank Srikant, Maya Taliaferro, Mingye Wang,
  Martin Schrimpf, Kendrick Kay, and Evelina Fedorenko.
\newblock Driving and suppressing the human language network using large
  language models.
\newblock \emph{bioRxiv}, 2023.

\bibitem[Wehbe et~al.(2014)Wehbe, Vaswani, Knight, and
  Mitchell]{wehbe_aligning_2014}
Leila Wehbe, Ashish Vaswani, Kevin Knight, and Tom Mitchell.
\newblock Aligning context-based statistical models of language with brain
  activity during reading.
\newblock In \emph{Proceedings of the 2014 {Conference} on {Empirical}
  {Methods} in {Natural} {Language} {Processing} ({EMNLP})}, pp.\  233--243,
  Doha, Qatar, October 2014. Association for Computational Linguistics.
\newblock \doi{10.3115/v1/D14-1030}.
\newblock URL \url{https://aclanthology.org/D14-1030}.

\bibitem[Wu et~al.(2006)Wu, David, and Gallant]{wu_complete_2006}
Michael C.-K. Wu, Stephen~V. David, and Jack~L. Gallant.
\newblock Complete functional characterization of sensory neurons by system
  identification.
\newblock \emph{Annual Review of Neuroscience}, 29:\penalty0 477--505, 2006.
\newblock ISSN 0147-006X.
\newblock \doi{10.1146/annurev.neuro.29.051605.113024}.

\bibitem[Yun et~al.(2021)Yun, Chen, Olshausen, and LeCun]{yun2021transformer}
Zeyu Yun, Yubei Chen, Bruno~A Olshausen, and Yann LeCun.
\newblock Transformer visualization via dictionary learning: contextualized
  embedding as a linear superposition of transformer factors.
\newblock \emph{arXiv preprint arXiv:2103.15949}, 2021.

\bibitem[Zaidan \& Eisner(2008)Zaidan and Eisner]{zaidan2008modeling}
Omar Zaidan and Jason Eisner.
\newblock Modeling annotators: A generative approach to learning from annotator
  rationales.
\newblock In \emph{Proceedings of the 2008 conference on Empirical methods in
  natural language processing}, pp.\  31--40, 2008.

\bibitem[Zeiler \& Fergus(2014)Zeiler and Fergus]{zeiler2014visualizing}
Matthew~D Zeiler and Rob Fergus.
\newblock Visualizing and understanding convolutional networks.
\newblock In \emph{European conference on computer vision}, pp.\  818--833.
  Springer, 2014.

\bibitem[Zellers et~al.(2019)Zellers, Bisk, Farhadi, and
  Choi]{zellers2019recognition}
Rowan Zellers, Yonatan Bisk, Ali Farhadi, and Yejin Choi.
\newblock From recognition to cognition: Visual commonsense reasoning.
\newblock In \emph{Proceedings of the IEEE/CVF conference on computer vision
  and pattern recognition}, pp.\  6720--6731, 2019.

\bibitem[Zhang et~al.(2023)Zhang, Xiao, Liu, Dou, and Nie]{zhang2023retrieve}
Peitian Zhang, Shitao Xiao, Zheng Liu, Zhicheng Dou, and Jian-Yun Nie.
\newblock Retrieve anything to augment large language models.
\newblock \emph{arXiv preprint arXiv:2310.07554}, 2023.

\bibitem[Zhang et~al.(2019)Zhang, Kishore, Wu, Weinberger, and
  Artzi]{zhang2019bertscore}
Tianyi Zhang, Varsha Kishore, Felix Wu, Kilian~Q Weinberger, and Yoav Artzi.
\newblock Bertscore: Evaluating text generation with bert.
\newblock \emph{arXiv preprint arXiv:1904.09675}, 2019.

\bibitem[Zhang et~al.(2015)Zhang, Zhao, and LeCun]{Zhang2015CharacterlevelCN}
Xiang Zhang, Junbo~Jake Zhao, and Yann LeCun.
\newblock Character-level convolutional networks for text classification.
\newblock In \emph{NIPS}, 2015.

\bibitem[Zhong et~al.(2021)Zhong, Lee, Zhang, and Klein]{zhong2021adapting}
Ruiqi Zhong, Kristy Lee, Zheng Zhang, and Dan Klein.
\newblock Adapting language models for zero-shot learning by meta-tuning on
  dataset and prompt collections.
\newblock \emph{arXiv preprint arXiv:2104.04670}, 2021.

\bibitem[Zhong et~al.(2022)Zhong, Snell, Klein, and
  Steinhardt]{zhong2022describing}
Ruiqi Zhong, Charlie Snell, Dan Klein, and Jacob Steinhardt.
\newblock Describing differences between text distributions with natural
  language.
\newblock In \emph{International Conference on Machine Learning}, pp.\
  27099--27116. PMLR, 2022.

\bibitem[Zhong et~al.(2023)Zhong, Zhang, Li, Ahn, Klein, and
  Steinhardt]{zhong2023goaldd}
Ruiqi Zhong, Peter Zhang, Steve Li, Jinwoo Ahn, Dan Klein, and Jacob
  Steinhardt.
\newblock Goal driven discovery of distributional differences via language
  descriptions.
\newblock \emph{arXiv preprint arXiv:2302.14233}, 2023.

\bibitem[Zhu et~al.(2022)Zhu, Liang, and Zou]{zhu2022gsclip}
Zhiying Zhu, Weixin Liang, and James Zou.
\newblock Gsclip: A framework for explaining distribution shifts in natural
  language.
\newblock \emph{arXiv preprint arXiv:2206.15007}, 2022.

\bibitem[Ziems et~al.(2023)Ziems, Held, Shaikh, Chen, Zhang, and
  Yang]{ziems2023can}
Caleb Ziems, William Held, Omar Shaikh, Jiaao Chen, Zhehao Zhang, and Diyi
  Yang.
\newblock Can large language models transform computational social science?
\newblock \emph{arXiv preprint arXiv:2305.03514}, 2023.

\end{thebibliography}
    \bibliographystyle{iclr2024_conference}
}

\appendix
\counterwithin{figure}{section}
\counterwithin{table}{section}
\renewcommand{\thefigure}{A\arabic{figure}}
\renewcommand{\thetable}{A\arabic{table}}

\section{Appendix}
\FloatBarrier

\subsection{Methodology details extended}

\begin{table}[h]
    \centering
    \caption{Statistics on corpuses used for explanation. Wikitext is used for BERT explanation and Moth stories are used for fMRI voxel explanation.}
    \begin{tabular}{lccc}
        \toprule
         & Unique unigrams & Unique bigrams & Unique trigrams \\
         \midrule
         Wikitext~\citep{merity2016pointer} & 157k & 3,719k & 9,228k\\
         Moth stories~\citep{lebel2022natural} & 117k & 79k & 140k\\
         Combined & 158k & 3,750k & 9,334k\\
         \bottomrule
    \end{tabular}
    \label{tab:dataset_statistics}
\end{table}

\paragraph{Prompts used in \methods}
The summarization step summarizes 30 randomly chosen ngrams from the top 50 and generates 5 candidate explanations using the prompt \textit{Here is a list of phrases:\textbackslash n\textcolor{darkgray}{\{phrases\}}\textbackslash nWhat is a common theme among these phrases?\textbackslash nThe common theme among these phrases is \blank}.

In the synthetic scoring step, we generate similar synthetic strings with the prompt \textit{Generate 10 phrases that are similar to the concept of \textcolor{darkgray}{\{explanation\}}:}.
For dissimilar synthetic strings we use the prompt \textit{Generate 10 phrases that are not similar to the concept of \textcolor{darkgray}{\{explanation\}}:}.
Minor automatic processing is applied to LLM outputs, e.g. parsing a bulleted list, converting to lowercase, and removing extra whitespaces.

\subsection{Synthetic module interpretation}
\label{subsec:synthetic_appendix}
\FloatBarrier

\begin{figure}[h]
    \centering
    \includegraphics[width=0.4\textwidth]{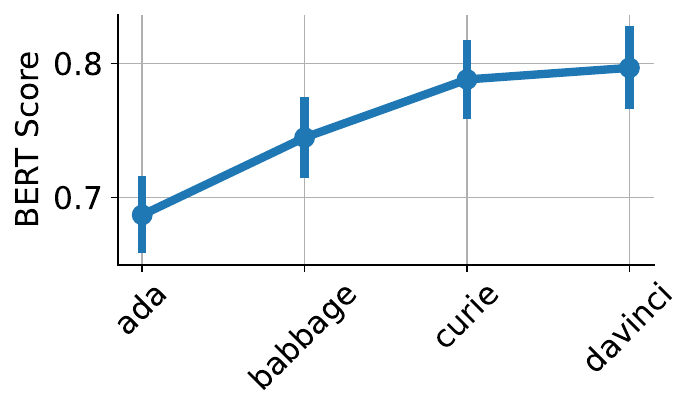}
    \caption{The BERT score between generated explanation and groundtruth explanation generally increases as the size of the helper LLM for summarization/generation increases.
    Models are accessed via the OpenAI API (\texttt{text-ada-001}, \texttt{text-babbage-001}, \texttt{text-curie-001}, \texttt{text-davinci-001}, all accessed on Feb. 2023) and are in order of increasing size.
    BERT score for each module is computed as the maximum over the 5 generated explanations.
    }
    \label{fig:synth_vary_llm}
\end{figure}

\begin{figure}[h]
    \centering
    \includegraphics[width=0.6\textwidth]{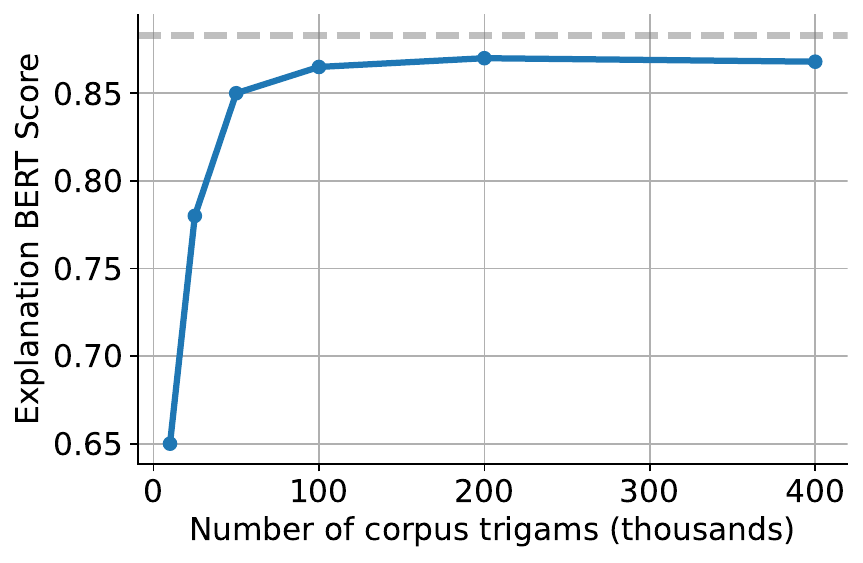}
    \caption{Explanation BERT score for the 54 synthetic datasets as a function of corpus size.
    Performance plateaus around 100,000 ngrams.
    Corpus is created by randomly subsampling the unique trigrams in the WikiText dataset~\citep{merity2016pointer}.
    Gray dotted line shows the result when evaluating on dataset-specific corpuses, as in the \textit{Default} setting in \cref{tab:recovery_results}.
    }
    \label{fig:synth_corpus_ablation}
\end{figure}

\begin{table}[h]
    \centering
    \scriptsize
    \caption{54 synthetic modules and information about their underlying data corpus. Note that some modules use the same groundtruth Keyword (e.g. \textit{environmentalism}), but that the underlying data corpus contains different data (e.g. text that is pro/anti environmentalism).}
    \begin{tabular}{ll p{0.35\textwidth}rr}
\toprule
Module name & Groundtruth keyphrase & Dataset explanation & Examples & Unique unigrams \\
\midrule
0-irony & sarcasm & contains irony & 590 & 3897 \\
1-objective & unbiased & is a more objective description of what happened & 739 & 5628 \\
2-subjective & subjective & contains subjective opinion & 757 & 5769 \\
3-god & religious & believes in god & 164 & 1455 \\
4-atheism & atheistic & is against religion & 172 & 1472 \\
5-evacuate & evacuation & involves a need for people to evacuate & 2670 & 16505 \\
6-terorrism & terrorism & describes a situation that involves terrorism & 2640 & 16608 \\
7-crime & crime & involves crime & 2621 & 16333 \\
8-shelter & shelter & describes a situation where people need shelter & 2620 & 16347 \\
9-food & hunger & is related to food security & 2642 & 16276 \\
10-infrastructure & infrastructure & is related to infrastructure & 2664 & 16548 \\
11-regime change & regime change & describes a regime change & 2670 & 16382 \\
12-medical & health & is related to a medical situation & 2675 & 16223 \\
13-water & water & involves a situation where people need clean water & 2619 & 16135 \\
14-search & rescue & involves a search/rescue situation & 2628 & 16131 \\
15-utility & utility & expresses need for utility, energy or sanitation & 2640 & 16249 \\
16-hillary & Hillary & is against Hillary & 224 & 1693 \\
17-hillary & Hillary & supports hillary & 218 & 1675 \\
18-offensive & derogatory & contains offensive content & 652 & 6109 \\
19-offensive & toxic & insult women or immigrants & 2188 & 11839 \\
20-pro-life & pro-life & is pro-life & 213 & 1633 \\
21-pro-choice & abortion & supports abortion & 209 & 1593 \\
22-physics & physics & is about physics & 10360 & 93810 \\
23-computer science & computers & is related to computer science & 10441 & 93947 \\
24-statistics & statistics & is about statistics & 9286 & 86874 \\
25-math & math & is about math research & 8898 & 85118 \\
26-grammar & ungrammatical & is ungrammatical & 834 & 2217 \\
27-grammar & grammatical & is grammatical & 826 & 2236 \\
28-sexis & sexist & is offensive to women & 209 & 1641 \\
29-sexis & feminism & supports feminism & 215 & 1710 \\
30-news & world & is about world news & 5778 & 13023 \\
31-sports & sports news & is about sports news & 5674 & 12849 \\
32-business & business & is related to business & 5699 & 12913 \\
33-tech & technology & is related to technology & 5727 & 12927 \\
34-bad & negative & contains a bad movie review & 357 & 16889 \\
35-good & good & thinks the movie is good & 380 & 17497 \\
36-quantity & quantity & asks for a quantity & 1901 & 5144 \\
37-location & location & asks about a location & 1925 & 5236 \\
38-person & person & asks about a person & 1848 & 5014 \\
39-entity & entity & asks about an entity & 1896 & 5180 \\
40-abbrevation & abbreviation & asks about an abbreviation & 1839 & 5045 \\
41-defin & definition & contains a definition & 651 & 4508 \\
42-environment & environmentalism & is against environmentalist & 124 & 1117 \\
43-environment & environmentalism & is environmentalist & 119 & 1072 \\
44-spam & spam & is a spam & 360 & 2470 \\
45-fact & facts & asks for factual information & 704 & 11449 \\
46-opinion & opinion & asks for an opinion & 719 & 11709 \\
47-math & science & is related to math and science & 7514 & 53973 \\
48-health & health & is related to health & 7485 & 53986 \\
49-computer & computers & related to computer or internet & 7486 & 54256 \\
50-sport & sports & is related to sports & 7505 & 54718 \\
51-entertainment & entertainment & is about entertainment & 7461 & 53573 \\
52-family & relationships & is about family and relationships & 7438 & 54680 \\
53-politic & politics & is related to politics or government & 7410 & 53393 \\
\bottomrule
\end{tabular}

    \label{tab:synthetic_examples_full}
\end{table}

\begin{table}[h]
    \centering
    \scriptsize
    \caption{54 synthetic datasets and the regex used to check whether an explanation is correct (after applying lowercasing). These regexes form guide the manual inspection of explanation accuracy: the original label is assigned by the regex and then fixed by the human when errors (which are relatively rare) occur.}
    \makebox[\textwidth][c]{
        \begin{tabular}{lll}
\toprule
Module name & Dataset explanation & Regex check \\
\midrule
0-irony & contains irony & irony$|$sarcas \\
1-objective & is a more objective description of what happened & objective$|$factual$|$nonpersonal$|$neutral$|$unbias \\
2-subjective & contains subjective opinion & subjective$|$opinion$|$personal$|$bias \\
3-god & believes in god & god$|$religious$|$religion \\
4-atheism & is against religion & atheism$|$atheist$|$anti-religion$|$against religion \\
5-evacuate & involves a need for people to evacuate & evacuat$|$flee$|$escape \\
6-terorrism & describes a situation that involves terrorism & terorrism$|$terror \\
7-crime & involves crime & crime$|$criminal$|$criminality \\
8-shelter & describes a situation where people need shelter & shelter$|$home$|$house \\
9-food & is related to food security & food$|$hunger$|$needs \\
10-infrastructure & is related to infrastructure & infrastructure$|$buildings$|$roads$|$bridges$|$build \\
11-regime change & describes a regime change & regime change$|$coup$|$revolution$|$revolt$|$political action$|$political event$|$upheaval \\
12-medical & is related to a medical situation & medical$|$health \\
13-water & involves a situation where people need clean water & water \\
14-search & involves a search/rescue situation & search$|$rescue$|$help \\
15-utility & expresses need for utility, energy or sanitation & utility$|$energy$|$sanitation$|$electricity$|$power \\
16-hillary & is against Hillary & hillary$|$clinton$|$against Hillary$|$opposed to Hillary$|$republican$|$against Clinton$|$opposed to Clinton \\
17-hillary & supports hillary & hillary$|$clinton$|$support Hillary$|$support Clinton$|$democrat \\
18-offensive & contains offensive content & offensive$|$toxic$|$abusive$|$insulting$|$insult$|$abuse$|$offend$|$offend$|$derogatory \\
19-offensive & insult women or immigrants & offensive$|$toxic$|$abusive$|$insulting$|$insult$|$abuse$|$offend$|$offend$|$women$|$immigrants \\
20-pro-life & is pro-life & pro-life$|$abortion$|$pro life \\
21-pro-choice & supports abortion & pro-choice$|$abortion$|$pro choice \\
22-physics & is about physics & physics \\
23-computer science & is related to computer science & computer science$|$computer$|$artificial intelligence$|$ai \\
24-statistics & is about statistics & statistics$|$stat$|$probability \\
25-math & is about math research & math$|$arithmetic$|$algebra$|$geometry \\
26-grammar & is ungrammatical & grammar$|$syntax$|$punctuation$|$grammat$|$linguistic \\
27-grammar & is grammatical & grammar$|$syntax$|$punctuation$|$grammat$|$linguistic \\
28-sexis & is offensive to women & sexis$|$women$|$femini \\
29-sexis & supports feminism & sexis$|$women$|$femini \\
30-news & is about world news & world$|$cosmopolitan$|$international$|$global \\
31-sports & is about sports news & sports \\
32-business & is related to business & business$|$economics$|$finance \\
33-tech & is related to technology & tech \\
34-bad & contains a bad movie review & bad$|$negative$|$awful$|$terrible$|$horrible$|$poor$|$boring$|$dislike \\
35-good & thinks the movie is good & good$|$great$|$like$|$love$|$positive$|$awesome$|$amazing$|$excellent \\
36-quantity & asks for a quantity & quantity$|$number$|$numeric \\
37-location & asks about a location & location$|$place \\
38-person & asks about a person & person$|$individual$|$people \\
39-entity & asks about an entity & entity$|$thing$|$object \\
40-abbrevation & asks about an abbreviation & abbrevation$|$abbr$|$acronym \\
41-defin & contains a definition & defin$|$meaning$|$explain \\
42-environment & is against environmentalist & environment$|$climate change$|$global warming \\
43-environment & is environmentalist & environment$|$climate change$|$global warming \\
44-spam & is a spam & spam$|$annoying$|$unwanted \\
45-fact & asks for factual information & fact$|$info$|$knowledge \\
46-opinion & asks for an opinion & opinion$|$personal$|$bias \\
47-math & is related to math and science & math$|$science \\
48-health & is related to health & health$|$medical$|$disease \\
49-computer & related to computer or internet & computer$|$internet$|$web \\
50-sport & is related to sports & sport \\
51-entertainment & is about entertainment & entertainment$|$music$|$movie$|$tv \\
52-family & is about family and relationships & family$|$relationships \\
53-politic & is related to politics or government & politic$|$government$|$law \\
\bottomrule
\end{tabular}

    }
    \label{tab:synthetic_regexes}
\end{table}

\begin{figure}[h]
    \centering
    \includegraphics[width=0.92\textwidth]{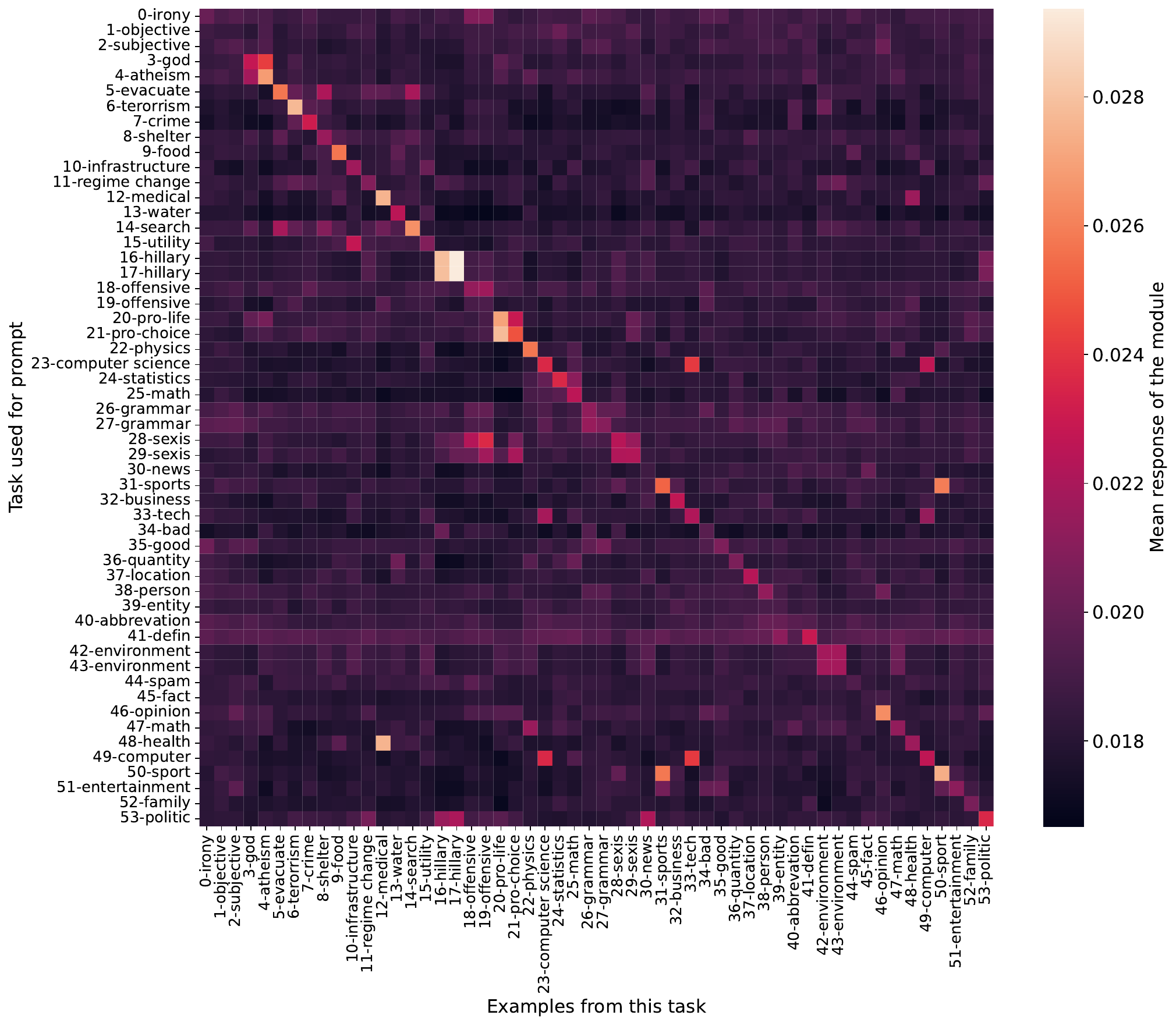}
    \caption{Synthetic modules respond more strongly to phrases related to their keyphrase (diagonal) than to phrases related to the keyphrase of other datasets (off-diagonal).
    Each value shows the mean response of the module to 5 phrases and each row is normalized using softmax.
    Each module is constructed using Instructor~\citep{su2022one} with the prompt \textit{Represent the short phrase for clustering: } and the groundtruth keyphrase given in \cref{tab:synthetic_examples_full}.
    Related keyphrases are generated manually.
    }
    \label{fig:mean_preds_synthetic}
\end{figure}

\begin{figure}[h]
    \centering
    \includegraphics[width=0.6\textwidth]{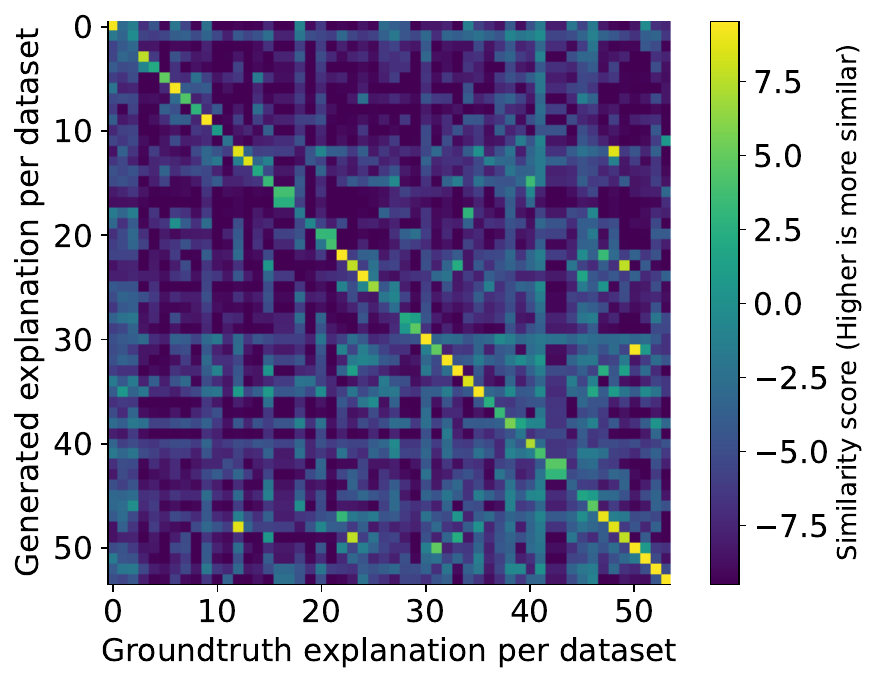}
    \caption{Similarity scores for SASC explanations in the \textit{Default} setting measured by bge-large (\texttt{BAAI/bge-large-en}, \citep{zhang2023retrieve}), rather than manual inspection or BERT-score, as shown in \cref{tab:recovery_results}. Large values on the diagonal indicate that the explanation generated for a module on a given dataset are similar to the groundtruth explanations for that dataset.
    The top-1 classification accuracy (i.e. how often the generated explanation is most similar to its corresponding groundtruth explanation) is 81.5\%, slightly lower than the assigned accuracy by manual inspection (88.3\%).
    The top-2 accuracy is 88.9\%.
    }
    \label{fig:synthetic_scores_sim}
\end{figure}

\begin{figure}[h]
    \centering
    \includegraphics[width=1.0\textwidth]{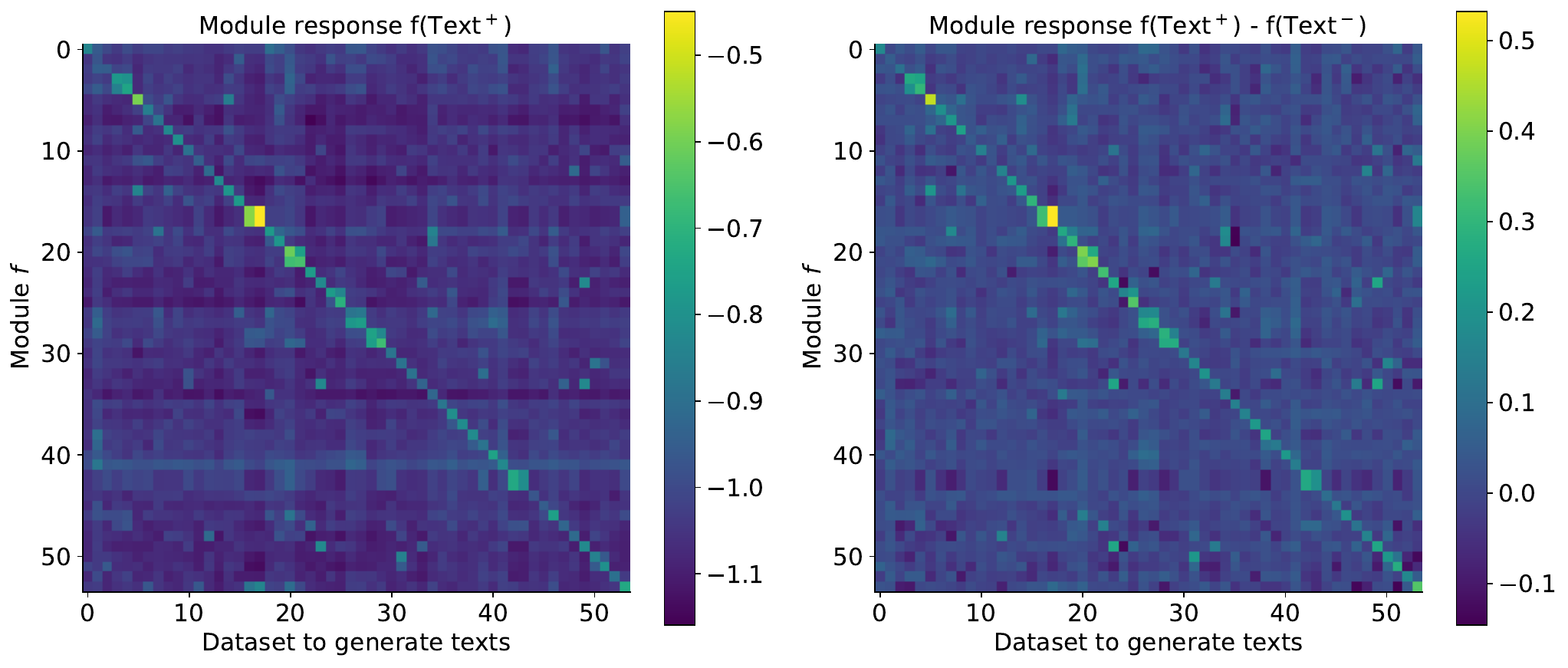}
    \caption{Average module responses for synthetic texts that are related to the explanation (left, $f(\text{Text}^+$)) or the difference between the responses for related and unrelated texts (right, $f(\text{Text}^+ - f(\text{Text}^-$)).
    Responses correspond to synthetic modules in the \textit{Default} setting.
    Bright diagonal on the left suggests that $f$ selectively responses to synthetic texts generated according to the appropriate explanation.
    On the right, the diagonal is slightly less bright, suggesting that the module does not tend to respond more negatively to unrelated texts $\text{Text}^-$.
    }
    \label{fig:mod_response_synthetic}
\end{figure}

\FloatBarrier
\subsection{BERT interpretation}
\label{subsec:transformer_factors_appendix}

\paragraph{Details on fitting transformer factors}
Pre-trained transformer factors are taken from ~\citep{yun2021transformer}.
Each transformer factor is the result of running dictionary learning on a matrix $X$ described as follows. Using a corpus of sentences $S$ (here wikipedia), embeddings are extracted for each input, layer, and sequence index in BERT. The resulting matrix $X$ has size $\left(\underbrace{\text{num\_layers}}_{\text{13 for BERT}} \cdot \sum_{s \in S} \text{len(s)}\right) \times \underbrace{d}_{\text{768 for BERT}}$.
Dictionary learning is run on $X$ with 1,500 dictionary components, resulting in a dictionary $D \in \mathbb R^{1,500 \times d}$.
Here, we take the fitted dictionary released by \citep{yun2021transformer} trained on the WikiText dataset~\citep{merity2016pointer}.

During our interpretation pipeline, we require a module which maps text to a scalar coefficient. To interpret a transformer factor as a module, we specify a text input $t$ and a layer $l$. This results in $len(t)$ embeddings with dimension $d$. We average over these embeddings, and then solve for the dictionary coefficients, to yield a set of coefficients $A \in \mathbb R^{1500}$. Finally, specifying a dictionary component index yields a single, scalar coefficient.

\paragraph{Extended BERT explanation results}
\cref{tab:bert_examples_full} shows examples comparing \methods explanations with human-labeled explanations for all BERT transformer factors labeled in \citep{yun2021transformer}.
\cref{tab:bert_sst,tab:bert_emotion,tab:bert_agnews} show explanations for modules selected by linear models finetuned on text-classification tasks.

\begin{table}[h]
    \centering
    \caption{Fraction of top logistic regression coefficients that are relevant for a downstream task (extends \cref{tab:bert_linear}). Averaged over 3 random seeds; parentheses show standard error of the mean.}
    \small
    \begin{tabular}{lccc}
        \toprule
         & Emotion & AG News & SST2 \\
         \midrule
         Top-10 & 0.50 \err{0.08} & 1.00 \err{0.00} & 0.80 \err{0.14}\\
         Top-15 & 0.47 \err{0.05} & 0.98 \err{0.03} & 0.69 \err{0.13}\\
         Top-20 & 0.42 \err{0.09} & 0.98 \err{0.02} & 0.55 \err{0.10}\\
         \bottomrule
    \end{tabular}
    \label{tab:bert_examples_full}
\end{table}

\begin{table}[h]
    \centering
    \caption{Comparing \methods explanations to all human-labeled explanations for BERT transformer factors. Explanation scores are in units of $\sigma_f$.}
    \scriptsize
    \renewcommand{\arraystretch}{1.2}
    \makebox[\textwidth][c]{
    \begin{tabular}{rrp{0.3\textwidth}p{0.3\textwidth}rr}
\toprule
\makecell{Factor\\Layer} & \makecell{Factor\\Index} & Explanation (Human) & Explanation (SASC) & \makecell{Explanation\\score\\(Human)} & \makecell{Explanation\\score\\(SASC)} \\
\midrule
4 & 13 & Numerical values. & numbers & -0.21 & -0.08 \\
10 & 42 & Something unfortunate happened. & idea of wrongdoing or illegal activity & 2.43 & 1.97 \\
0 & 30 & left. Adjective or Verb. Mixed senses. & someone or something leaving & 3.68 & 5.87 \\
4 & 47 & plants. Noun. vegetation. & trees & 6.26 & 5.04 \\
10 & 152 & In some locations. & science, technology, and/or medicine & -0.41 & 0.03 \\
4 & 30 & left. Verb. leaving, exiting. & leaving or being left & 4.44 & 0.90 \\
10 & 297 & Repetitive structure detector. & versions or translations & -0.36 & 0.98 \\
10 & 322 & Biography, someone born in some year... & weapons and warfare & 0.19 & 0.38 \\
10 & 13 & Unit exchange with parentheses. & names of places, people, or things & -0.11 & -0.10 \\
10 & 386 & War. & media, such as television, movies, or video games & 0.20 & -0.15 \\
10 & 184 & Institution with abbreviation. & publishing, media, or awards & -0.42 & 0.14 \\
2 & 30 & left. Verb. leaving, exiting. & leaving or being left & 5.30 & 0.91 \\
10 & 179 & Topic: music production. & geography & -0.52 & 0.21 \\
6 & 225 & Places in US, followings the convention "city, state". & a place or location & 1.88 & 1.33 \\
10 & 25 & Attributive Clauses. & something related to people, places, or things & 0.01 & 1.19 \\
10 & 125 & Describing someone in a para- phrasing style. Name, Career. & something related to buildings, architecture, or construction & -0.13 & 0.44 \\
6 & 13 & Close Parentheses. & end with a closing punctuation mark (e.g & -0.08 & 0.47 \\
10 & 99 & Past tense. & people, places, or things & -0.77 & -0.04 \\
10 & 24 & Male name. & people, places, and things related to history & 0.03 & 0.38 \\
10 & 102 & African names. & traditional culture, with references to traditional territories, communities, forms, themes, breakfast, and texts & 0.35 & 1.60 \\
4 & 16 & park. Noun. a common first and last name. & names of parks & -0.03 & 1.87 \\
10 & 134 & Transition sentence. & a comma & 1.16 & 0.38 \\
6 & 86 & Consecutive years, used in foodball season naming. & specific dates or months & 0.85 & 0.76 \\
4 & 2 & mind. Noun. the element of a person that enables them to be aware of the world and their experiences. & concept of thinking, remembering, and having memories & 0.77 & 11.19 \\
10 & 51 & Apostrophe s, possesive. & something specific, such as a ticket, tenure, film, song, movement, project, game, school, title, park, congressman, author, or art exhibition & 0.37 & -0.01 \\
8 & 125 & Describing someone in a paraphrasing style. Name, Career. & publications, reviews, or people associated with the media industry & -0.34 & 0.42 \\
4 & 33 & light. Noun. the natural agent that stimulates sight and makes things visible. & light & 6.25 & 3.43 \\
10 & 50 & Doing something again, or making something new again. & introduction of something new & 0.84 & -0.27 \\
10 & 86 & Consecutive years, this is convention to name foodball/rugby game season. & a specific date or time of year & 1.35 & -0.75 \\
4 & 193 & Time span in years. & many of them are related to dates and historic places & 0.07 & 1.39 \\
10 & 195 & Consecutive of noun (Enumerating). & different aspects of culture, such as art, music, literature, history, and technology & -0.83 & 9.83 \\
\bottomrule
\end{tabular}

    }
    \label{tab:bert_examples_full2}
\end{table}

\begin{table}[ht]
    \centering
    \scriptsize
    \caption{\methods explanations for modules selected by 25-coefficient linear model on \textit{SST2} for a single seed. Green shows explanations deemed to be relevant to the task.}
    \renewcommand{\arraystretch}{1.2}
    \makebox[\textwidth][c]{
    \begin{tabular}{llrll}
\toprule
Layer, Factor index & Explanation & Linear coefficient               &  \\
\midrule
(0, 783)                                     & \cellcolor[HTML]{D9EAD3}something being incorrect or wrong                                           & -862.82                                                                                        &  &  \\
(0, 1064)                                    & \cellcolor[HTML]{D9EAD3}negative emotions and actions, such as hatred, violence, and disgust         & -684.27                                                                                        &  &  \\
(1, 783)                                     & \cellcolor[HTML]{D9EAD3}something being incorrect, inaccurate, or wrong                              & -577.49                                                                                        &  &  \\
(1, 1064)                                    & \cellcolor[HTML]{D9EAD3}hatred and violence                                                          & -499.30                                                                                        &  &  \\
(0, 157)                                     & air and sequencing                                                                                   & 463.80                                                                                         &  &  \\
(9, 319)                                     & \cellcolor[HTML]{D9EAD3}a negative statement, usually in the form of not or nor                      & -446.58                                                                                        &  &  \\
(0, 481)                                     & \cellcolor[HTML]{D9EAD3}harm, injury, or damage                                                      & -441.98                                                                                        &  &  \\
(8, 319)                                     & lack of something or the absence of something                                                        & -441.04                                                                                        &  &  \\
(10, 667)                                    & two or more words                                                                                    & 424.48                                                                                         &  &  \\
(2, 783)                                     & \cellcolor[HTML]{D9EAD3}something that is incorrect or inaccurate                                    & -415.56                                                                                        &  &  \\
(0, 658)                                     & thrice                                                                                               & -411.26                                                                                        &  &  \\
(0, 319)                                     & \cellcolor[HTML]{D9EAD3}none or its variations (no, not, never)                                      & -388.14                                                                                        &  &  \\
(0, 1402)                                    & dates                                                                                                & -377.74                                                                                        &  &  \\
(0, 1049)                                    & standard                                                                                             & -365.83                                                                                        &  &  \\
(3, 1064)                                    & \cellcolor[HTML]{D9EAD3}negative emotions or feelings, such as hatred, anger, disgust, and brutality & -360.47                                                                                        &  &  \\
(4, 1064)                                    & \cellcolor[HTML]{D9EAD3}negative emotions or feelings, such as hatred, anger, and disgust            & -357.35                                                                                        &  &  \\
(5, 152)                                     & geography, history, and culture                                                                      & -356.10                                                                                        &  &  \\
(0, 928)                                     & homelessness and poverty                                                                             & -355.05                                                                                        &  &  \\
(2, 691)                                     & animals and plants, as many of the phrases refer to species of animals and plants                    & -351.62                                                                                        &  &  \\
(0, 810)                                     & catching or catching something                                                                       & 350.98                                                                                         &  &  \\
(0, 1120)                                    & production                                                                                           & -350.01                                                                                        &  &  \\
(0, 227)                                     & a period of time                                                                                     & -345.72                                                                                        &  &  \\
(2, 583)                                     & government, law, or politics in some way                                                             & -335.40                                                                                        &  &  \\
(2, 1064)                                    & \cellcolor[HTML]{D9EAD3}negative emotions such as hatred, disgust, and violence                      & -334.87                                                                                        &  &  \\
(4, 125)                                     & science or mathematics, such as physics, astronomy, and geometry                                     & -328.55                                                                                        &  & \\
\bottomrule
\end{tabular}
    }
    \label{tab:bert_sst}
\end{table}   

\begin{table}[ht]
    \centering
    \scriptsize
    \caption{\methods explanations for modules selected by 25-coefficient linear model on \textit{AG News} for a single seed. Green shows explanations deemed to be relevant to the task.}
    \renewcommand{\arraystretch}{1.2}
    \makebox[\textwidth][c]{
    \begin{tabular}{llrll}
\toprule
Layer, Factor index & Explanation & Linear coefficient              &  \\
\midrule
(5, 378)                                     & \cellcolor[HTML]{D9EAD3}professional sports teams                                                  & 545.57                                                                                         &               &  \\
(4, 378)                                     & \cellcolor[HTML]{D9EAD3}professional sports teams in the united states                             & 542.25                                                                                         &               &  \\
(3, 378)                                     & \cellcolor[HTML]{D9EAD3}professional sports teams                                                  & 515.37                                                                                         &               &  \\
(0, 378)                                     & \cellcolor[HTML]{D9EAD3}names of sports teams                                                      & 508.73                                                                                         &               &  \\
(6, 378)                                     & \cellcolor[HTML]{D9EAD3}sports teams                                                               & 499.62                                                                                         &  \\
(2, 378)                                     & \cellcolor[HTML]{D9EAD3}professional sports teams                                                  & 499.57                                                                                         &               &  \\
(1, 378)                                     & \cellcolor[HTML]{D9EAD3}professional sports teams                                                  & 492.01                                                                                         &               &  \\
(7, 378)                                     & \cellcolor[HTML]{D9EAD3}sports teams                                                               & 468.66                                                                                         &               &  \\
(8, 378)                                     & \cellcolor[HTML]{D9EAD3}sports teams or sports in some way                                         & 468.39                                                                                         &               &  \\
(11, 32)                                     & \cellcolor[HTML]{D9EAD3}activity or process                                                        & 461.46                                                                                         &               &  \\
(12, 1407)                                   & such                                                                                               & 450.70                                                                                         &               &  \\
(5, 730)                                     & \cellcolor[HTML]{D9EAD3}england and english sports teams                                           & 427.33                                                                                         &               &  \\
(12, 104)                                    & \cellcolor[HTML]{D9EAD3}people, places, and events from history                                    & 425.49                                                                                         &               &  \\
(10, 378)                                    & \cellcolor[HTML]{D9EAD3}locations                                                                  & 424.71                                                                                         &               &  \\
(6, 730)                                     & \cellcolor[HTML]{D9EAD3}sports, particularly soccer                                                & 424.24                                                                                         &               &  \\
(12, 730)                                    & \cellcolor[HTML]{D9EAD3}sports                                                                     & 415.21                                                                                         &               &  \\
(4, 396)                                     & \cellcolor[HTML]{D9EAD3}people, places, or things related to japan                                 & -415.13                                                                                        &               &  \\
(10, 659)                                    & \cellcolor[HTML]{D9EAD3}sports                                                                     & 410.89                                                                                         &               &  \\
(4, 188)                                     & \cellcolor[HTML]{D9EAD3}history in some way                                                        & 404.24                                                                                         &               &  \\
(12, 1465)                                   & \cellcolor[HTML]{D9EAD3}different aspects of life, such as activities, people, places, and objects & 403.77                                                                                         &               &  \\
(0, 310)                                     & end with the word until                                                                            & -400.10                                                                                        &               &  \\
(5, 151)                                     & \cellcolor[HTML]{D9EAD3}a particular season, either of a year, a sport, or a television show       & 396.41                                                                                         &               &  \\
(12, 573)                                    & many of them contain unknown words or names, indicated by \textless unk                            & -393.27                                                                                        &               &  \\
(12, 372)                                    & \cellcolor[HTML]{D9EAD3}specific things, such as places, organizations, or activities              & -392.57                                                                                        &               &  \\
(6, 188)                                     & \cellcolor[HTML]{D9EAD3}geography                                                                  & 388.69                                                                                         &               & \\
\bottomrule
\end{tabular}
    }
    \label{tab:bert_agnews}
\end{table}   

\begin{table}[ht]
    \centering
    \scriptsize
    \caption{\methods explanations for modules selected by 25-coefficient linear model on \textit{Emotion} for a single seed. Green shows explanations deemed to be relevant to the task.}
    \renewcommand{\arraystretch}{1.2}
    \makebox[\textwidth][c]{
    \begin{tabular}{llrll}
\toprule
Layer, Factor index & Explanation & Linear coefficient &               &  \\
\midrule
(0, 1418)                                    & \cellcolor[HTML]{D9EAD3}types of road interchanges                                                        & 581.97                                                                                         &               &  \\
(0, 920)                                     & \cellcolor[HTML]{D9EAD3}fame                                                                              & 577.20                                                                                         &               &  \\
(6, 481)                                     & \cellcolor[HTML]{D9EAD3}injury or impairment                                                              & 566.44                                                                                         &               &  \\
(5, 481)                                     & \cellcolor[HTML]{D9EAD3}injury or impairment                                                              & 556.58                                                                                         &               &  \\
(0, 693)                                     & \cellcolor[HTML]{D9EAD3}end in oss or osses                                                               & 556.53                                                                                         & &  \\
(12, 1137)                                   & \cellcolor[HTML]{D9EAD3}ownership or possession                                                           & -537.45                                                                                        &               &  \\
(0, 663)                                     & \cellcolor[HTML]{D9EAD3}civil                                                                             & 524.88                                                                                         &               &  \\
(6, 1064)                                    & \cellcolor[HTML]{D9EAD3}negative emotions such as hatred, disgust, disdain, rage, and horror              & 523.41                                                                                         &               &  \\
(3, 872)                                     & location of a campus or facility                                                                          & -518.85                                                                                        &               &  \\
(5, 1064)                                    & \cellcolor[HTML]{D9EAD3}negative emotions and feelings, such as hatred, disgust, disdain, and viciousness & 489.25                                                                                         &               &  \\
(0, 144)                                     & lectures                                                                                                  & 482.85                                                                                         &               &  \\
(0, 876)                                     & \cellcolor[HTML]{D9EAD3}host                                                                              & 479.18                                                                                         &               &  \\
(0, 69)                                      & history                                                                                                   & -467.80                                                                                        &               &  \\
(0, 600)                                     & many of them contain the word seymour or a variation of it                                                & 464.64                                                                                         &               &  \\
(0, 813)                                     & \cellcolor[HTML]{D9EAD3}or phrases related to either measurement (e.g                                     & -455.11                                                                                        &               &  \\
(1, 89)                                      & \cellcolor[HTML]{D9EAD3}caution and being careful                                                         & 451.73                                                                                         &               &  \\
(11, 229)                                    & russia and russian culture                                                                                & -450.28                                                                                        &               &  \\
(0, 783)                                     & \cellcolor[HTML]{D9EAD3}something being incorrect or wrong                                                & 448.55                                                                                         &               &  \\
(12, 195)                                    & dates                                                                                                     & 442.14                                                                                         &               &  \\
(12, 1445)                                   & \cellcolor[HTML]{D9EAD3}breaking or being broken                                                          & 439.81                                                                                         &               &  \\
(0, 415)                                     & ashore                                                                                                    & -438.22                                                                                        &               &  \\
(0, 118)                                     & end with a quotation mark                                                                                 & 437.66                                                                                         &               &  \\
(1, 650)                                     & mathematical symbols such as \textgreater{}, =, and )                                                     & -437.28                                                                                        &               &  \\
(4, 388)                                     & \cellcolor[HTML]{D9EAD3}end with the sound ch                                                             & -437.15                                                                                        &               &  \\
(0, 840)                                     & withdrawing                                                                                               & -436.38                                                                                        &               & \\
\bottomrule
\end{tabular}
    }
    \label{tab:bert_emotion}
\end{table}

\FloatBarrier
\subsection{fMRI module interpretation}
\label{subsec:fmri_appendix}

\subsubsection{fMRI data and model fitting}
\label{subsec:fmri_setup}
This section gives more details on the fMRI experiment analyzed in \cref{sec:fmri_results}. These MRI data are available publicly \citep{lebel2022natural,tang2023semantic}, but the methods are summarized here. Functional magnetic resonance imaging (fMRI) data were collected from 3 human subjects as they listened to English language podcast stories over Sensimetrics S14 headphones. Subjects were not asked to make any responses, but simply to listen attentively to the stories. For encoding model training, each subject listened to at approximately 20 hours of unique stories across 20 scanning sessions, yielding a total of  \(\sim \)33,000 datapoints for each voxel across the whole brain. For model testing, the subjects listened to two test story 5 times each, and one test story 10 times, at a rate of 1 test story per session. These test responses were averaged across repetitions. Functional signal-to-noise ratios in each voxel were computed using the mean-explainable variance method from \citep{nishimoto2017eye} on the repeated test data. Only voxels within 8 mm of the mid-cortical surface were analyzed, yielding roughly 90,000 voxels per subject. 

MRI data were collected on a 3T Siemens Skyra scanner at University of Texas at Austin using a 64-channel Siemens volume coil. Functional scans were collected using a gradient echo EPI sequence with repetition time (TR) = 2.00 s, echo time (TE) = 30.8 ms, flip angle = 71°, multi-band factor (simultaneous multi-slice) = 2, voxel size = 2.6mm x 2.6mm x 2.6mm (slice thickness = 2.6mm), matrix size = 84x84, and field of view = 220 mm. Anatomical data were collected using a T1-weighted multi-echo MP-RAGE sequence with voxel size = 1mm x 1mm x 1mm following the Freesurfer morphometry protocol ~\citep{fischl2012freesurfer}.

All subjects were healthy and had normal hearing. The experimental protocol was approved by the Institutional Review Board at the University of Texas at Austin. Written informed consent was obtained from all subjects.

All functional data were motion corrected using the FMRIB Linear Image Registration Tool (FLIRT) from FSL 5.0. FLIRT was used to align all data to a template that was made from the average across the first functional run in the first story session for each subject. These automatic alignments were manually checked for accuracy. 
 
Low frequency voxel response drift was identified using a 2nd order Savitzky-Golay filter with a 120 second window and then subtracted from the signal. To avoid onset artifacts and poor detrending performance near each end of the scan, responses were trimmed by removing 20 seconds (10 volumes) at the beginning and end of each scan, which removed the 10-second silent period and the first and last 10 seconds of each story. The mean response for each voxel was subtracted and the remaining response was scaled to have unit variance.

We used the fMRI data to generate a voxelwise brain encoding model for natural language using the intermediate hidden states from the 
the 18th layer of the 30-billion parameter LLaMA model~\citep{touvron2023llama},
and the 9th layer of GPT~\citep{radford2019language}.
In order to temporally align word times with TR times, Lanczos interpolation was applied with a window size of 3. The hemodyanmic response function was approximated with a finite impulse response model using 4 delays at -8,-6,-4 and -2 seconds \citep{huth2016natural}. For each subject $x$, voxel $v$, we fit a separate encoding model $g_{(x,v)}$ to predict the BOLD response $\hat{B}$ from our embedded stimulus, i.e. $\hat{B}_{(x,v)} = g_{(x,v)}(H_i(\set{S}))$. 

To evaluate the voxelwise encoding models, we used the learned $g_{(x,v)}$ to generate and evaluate predictions on a held-out test set.
The
GPT features achieved a mean correlation of 0.12 and
LLaMA features achieved a mean correlation of 0.17.
These performances are comparable with state-of-the-art published models on the same dataset that are able to achieved decoding~\citep{tang2023semantic}.

To select voxels with diverse encoding, we applied principal components analysis to the learned weights, $g_{(x,v)}$, for GPT across all significantly predicted voxels in cortex. Prior work has shown that the first four principal components of language encoding models weights encode differences in semantic selectivity, differentiating between concepts like \textit{social}, \textit{temporal} and \textit{visual} concepts. Consequently, to apply \methods to voxels with the most diverse selectivity, we found voxels that lie along the convex hull of the first four principal components and randomly sampled 1,500 of them (500 per subject).
The mean voxel correlation for the 1,500 voxels we study is 0.35.
Note that these voxels were selected for being well-predicted rather than easy to explain: the correlation between the prediction error and the explanation score for these voxels is 0.01, very close to zero.

\subsubsection{Evaluating top fMRI voxel evaluations}
\label{subsec:fmri_voxel_eval_supp}
\cref{tab:fmri_eval} shows two evaluations of the fMRI voxel explanations.
First, similar to \cref{fig:syn_perc_score_boxplot}, we find the mean explanation score remains significantly above zero.
Second, we evaluate beyond whether the explanation describes the fitted module and ask whether the explanation describes the underlying fMRI voxel.
Specifically, we predict the fMRI voxel response to text using only the voxel's explanation using a very simple procedure.
We first compute the (scalar) negative embedding distance between the explanation text and the input text using Instructor~\citep{su2022one}\footnote{The input text for an fMRI response at time $t$ (in seconds) is taken to be the words presented between $t - 8$ and $t - 2$.}.
We then calculate the spearman rank correlation between this scalar distance and the recorded voxel response (see \cref{tab:fmri_eval}).
The mean computed correlation is low\footnote{For reference, test correlations published in fMRI voxel prediction from language are often in the range of 0.01-0.1~\citep{caucheteux2022deep}.}, which is to be expected as the explanation is a concise string and may match extremely few ngrams in the text of the test data (which consists of only 3 narrative stories).
Nevertheless, the correlation is significantly above zero (more than 15 times the standard error of the mean), suggesting that these explanations have some grounding in the underlying brain voxels.

\begin{table}[ht]
    \centering
    \caption{
    Evaluation of fMRI voxel explanations.
    For all metrics, \methods is successful if the value is significantly greater than 0. Errors show standard error of the mean.}
    \small
    \begin{tabular}{ll}
    \toprule
    Explanation score & Test rank correlation \\
    \midrule
    1.27$\sigma_f$ \err{0.029} & 0.033 \err{0.002} \\
    \bottomrule
    \end{tabular}
    \label{tab:fmri_eval}
\end{table}

\subsubsection{fMRI results when using WikiText Corpus}
\label{subsec:fmri_wikitext}

\begin{figure}[h]
    \centering
    \includegraphics[width=0.65\textwidth]{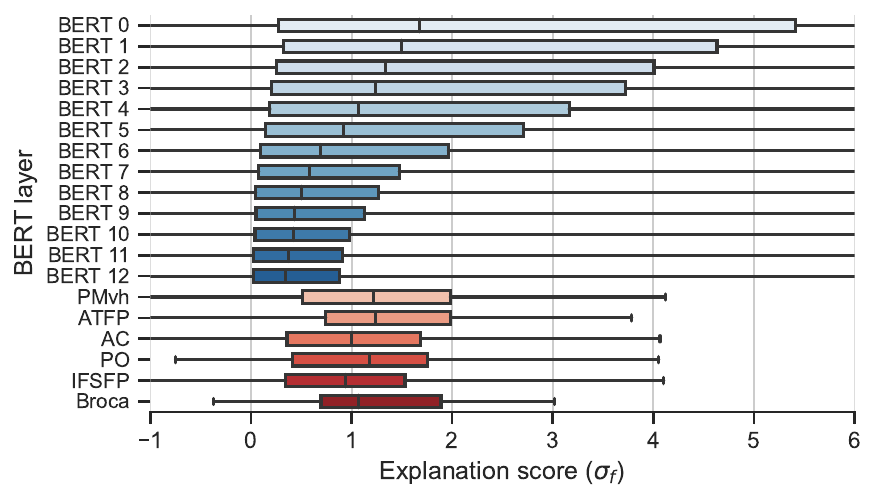}
    \vspace{-12pt}
    \caption{Results in \cref{fig:syn_perc_score_boxplot} when using WikiText as the underlying corpus for ngrams rather than narrative stories.}
    \label{fig:syn_perc_wiki}
\end{figure}

\begin{figure}[h]
    \centering
    \includegraphics[width=0.9\textwidth]{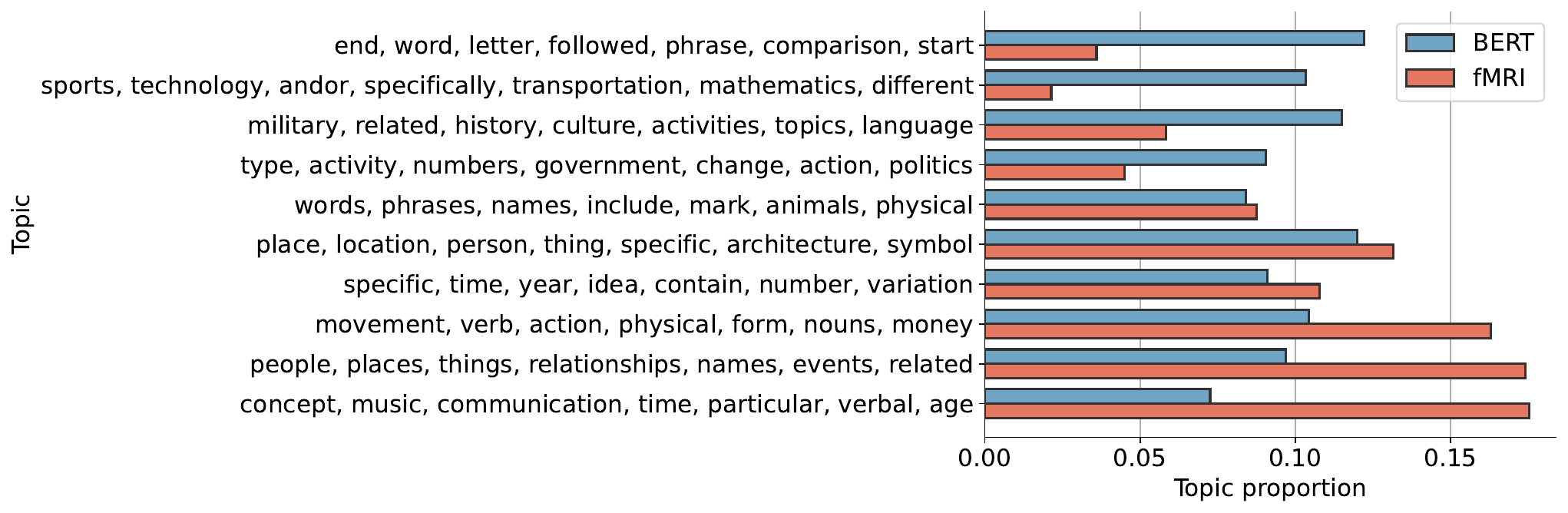}
    \vspace{-12pt}
    \caption{Results in \cref{fig:topic_proportions} when using WikiText as the underlying corpus for ngrams rather than narrative stories.}
    \label{fig:topic_proprtions_wiki}
\end{figure}

\end{document}